\documentclass[accepted]{uai2022} % after acceptance, for a revised
\usepackage[american]{babel}
\usepackage{natbib} % has a nice set of citation styles and commands
\usepackage{mathtools} % amsmath with fixes and additions
\usepackage{booktabs} % commands to create good-looking tables
\usepackage{tikz} % nice language for creating drawings and diagrams
\usepackage{todonotes}

\title{Offline Policy Optimization with Eligible Actions}

\author[1]{Yao Liu\thanks{This work is mostly done when the author was at Stanford.}}
\author[2]{Yannis Flet-Berliac}
\author[2]{Emma Brunskill}

\affil[1]{%
    ByteDance\\
    yao.liu.chn@gmail.com
}
\affil[2]{%
    Stanford University\\
    \{yfletberliac, ebrun\}@cs.stanford.edu
}

\usepackage{hyperref}       % hyperlinks
\usepackage{url}            % simple URL typesetting
\usepackage{booktabs}       % professional-quality tables
\usepackage{amsfonts}       % blackboard math symbols
\usepackage{nicefrac}       % compact symbols for 1/2, etc.
\usepackage{microtype}      % microtypography
\usepackage{xcolor}         % colors
\usepackage{wrapfig}
\usepackage{natbib}

\usepackage{my_definitions}
\newcommand{\transitiondescriptor}{P}
\newcommand{\transitionkernel}[1]{P_{#1}}
\newcommand{\rewarddescriptor}{R}
\newcommand{\rewardkernel}[1]{R_{#1}}

\newcommand{\ith}[1]{#1^{(i)}}
\newcommand{\Wsum}{W}

\newcommand{\wtis}{\hat{v}_{\text{SNTIS}}}
\newcommand{\Dcal}{\mathcal{D}}
\newcommand{\Bcal}{\mathcal{B}}
\newcommand{\empE}{\hat{\EE}}

\newcommand{\Aset}{A}
\newcommand{\dist}{\mathrm{dist}}
\newcommand{\threshold}{\delta}

\newcommand{\Prob}{\mathrm{P}}
\newcommand{\ispg}{\textsc{POELA}\xspace}
\newcommand{\ispgbaseline}{PO-CRM\xspace}
\newcommand{\ispgknn}{PO-$\mu$}
\newcommand{\ispgmuhat}{PO-$\hat{\mu}$}

\newtheorem{example}{Example}

\usepackage[
  separate-uncertainty = true,
  multi-part-units = repeat
]{siunitx}

\begin{document}

\maketitle

\begin{abstract}
Offline policy optimization could have a large impact on many real-world decision-making problems, as online learning may be infeasible in many applications. Importance sampling and its variants are a commonly used type of estimator in offline policy evaluation, and such estimators typically do not require assumptions on the properties and representational capabilities of value function or decision process model function classes. In this paper, we identify an important overfitting phenomenon in optimizing the importance weighted return, in which it may be possible for the learned policy to essentially avoid making aligned decisions for part of the initial state space. We propose an algorithm to avoid this overfitting through a new per-state-neighborhood normalization constraint, and provide a theoretical justification of the proposed algorithm. We also show the limitations of previous attempts to this approach. We test our algorithm in a healthcare-inspired simulator, a logged dataset collected from real hospitals and continuous control tasks. These experiments show the proposed method yields less overfitting and better test performance compared to state-of-the-art batch reinforcement learning algorithms.
\end{abstract}

\section{Introduction}
\label{sec:intro}
There has been a recent surge in interest, from both the theoretical and algorithmic perspective, in offline/batch Reinforcement Learning (RL). This area could potentially bring insights from RL to the growing number of application settings which produce such datasets (like healthcare~\citep{gottesman2019guidelines,nie2021learning}, customer marketing~\citep{bottou2013counterfactual,thomas2017predictive} or home automation~\citep{emmons2022the}), provide ways to leverage vast amounts of observational training data encoded in videos~\citep{chang2020semantic,pmlr-v155-schmeckpeper21a}% (consider youtube as a training data set for training robots to cook, folder laundry or entertain at parties)
, or advance our core understanding of the data characteristics needed to learn near optimal policies. 

Many of the settings that might benefit most from offline RL, like healthcare, education or autonomous driving, may not be Markov in the available observed features, and also may not include explicit known representations of the behavior decision policy. This has inspired work on offline policy evaluation estimation methods that make minimal assumptions on the data generation process, such as importance sampling (IS) and doubly robust estimation methods~\citep{precup2000eligibility,Jiang2016,thomas2016data} and offline policy learning methods that leverage such estimators~\citep{huang2020importance,cheng2020trajectory,thomas2019preventing}.

Unfortunately, we show that offline policy selection or learning algorithms that rely on such offline estimation methods that leverage IS can suffer from a key flaw. In brief, the structure of the policy estimator is such that high estimated performance can be achieved both by policies that have high average performance, and by policies that systematically avoid taking actions taken in the dataset for initial states that lead to low rewards. This can substantially inflate the estimated performance of a potential decision policy. As an intuition, consider a setting where some patients arrive much sicker than some healthier patients. In this setting, any policy for the sicker patients will likely yield slightly worse outcomes than the average outcomes for healthier patients, but a policy’s value must be taken in expectation over all patients, not just the healthy patients. We detail how a number of methods, including those that have been proposed before for other reasons, do not solve this issue, including: using a validation set, shifting the reward baseline, and leveraging thresholds on an estimated behavior policy. 

Fortunately, we show that a relatively simple method for constraining the policy class considered with off policy learning can greatly ameliorate the problem of propensity overfitting. Our approach can be viewed as related to pessimistic offline RL in Markov decision processes, which has relations to the robust MDP literature~\citep{nilim2005robust}
%(**cite drew, Berkeley people, marek? Etc], was revived recently~\citep{DBLP:conf/uai/LiuSAB19,satija2021multiobjective} 
and has been receiving growing attention (e.g.,~\cite{liu2020provably,yu2020mopo,kidambi2020morel}).
%with a number of significant empirical improvements (** cite a bunch) and theoretical advances (**cite) in continuous state and/or action settings and in conjunction with deep RL. 
One of the key tenants of pessimistic offline RL is to maintain precise quantification over the uncertainty over the model parameters and/or value functions of the Markov decision process, given the available data.  A key challenge is how to quantify statistical uncertainty when the state space is extremely large or continuous. This issue is perhaps even more paramount in offline RL settings when we wish to leverage IS-based estimators in order to make minimal assumptions over the data generation process. We show here that constraining the policy class per state only to actions taken in the data for nearby states, which may be viewed as a loose analogy to count-based uncertainty, is sufficient to lower bound the amount of propensity overfitting that can occur. Our approach still ensures asymptotic consistency of the estimation of any policy covered by the behavior policy\footnote{Coverage is a minimal requirement for all IS methods to be consistent estimators of a new proposed policy.} while providing significant benefits in the finite regime, by essentially constraining policies to observed actions. In this way, our method is related to other methods that revert back to the behavior policy given minimal data for MDP model or value function learning~\citep{satija2021multiobjective} or in the bandit setting~\citep{sachdeva2020off,brandfonbrener2020offline}. To our knowledge, our work is the first to explore this for large, non-Markovian stochastic decision processes using policy search methods.
%and also the first to consider how to do this in very large state spaces when the behavior policy is unknown or when the state and action space is sufficiently large that careful attention to being precise about what it means for a state-action pair to be sufficiently well estimated is paramount. [**revise this to be precise. Spibb did look at large spaces but not sure thought about how to regularize back to $pi_b$ in such spaces and if $pi_b$ is unknown]. 
We show in simulations and in a real dataset on patient sepsis outcomes that our approach learns policies with significantly higher expected rewards than prior methods, and that those estimates are expected to be reliable, with solid effective samples sizes, a measure of how much of the behavior data would match the proposed policy. Our results highlight and remedy a potential reliability barrier for offline RL with minimal data process and realizability assumptions.
%and suggest interesting open issues about the role of quantifying uncertainty in offline datasets. 

\section{Offline Policy Optimization}
\label{sec:problem_setting}
We study the problem of offline policy optimization in sequential decision-making under uncertainty. Let the environment be a finite-horizon Contextual Decision Process (CDP) \citep{jiang2017contextual}. A CDP can capture more general, non-Markovian settings (also sometimes referred to as a Non-Markov DP \citep{kallus2019efficiently}). A CDP is 
defined as a tuple $\langle \Xcal, \Acal, H, \transitiondescriptor, \rewarddescriptor \rangle$, where $\Xcal$ is the context space, $\Acal$ is the action space, and $H$ is the horizon. $\transitiondescriptor = \{  \transitionkernel{h} \}_{h=1}^H$ is the unknown transition model, where $\transitionkernel{h}: (\Xcal \times \Acal)^{h-1} \to \Delta(\Xcal)$ is the distribution over next context given the history. $\transitionkernel{1}: \Delta(\Xcal)$ is the initial context distribution. Similarly, $\rewarddescriptor = \{  \rewardkernel{h} \}_{h=1}^H$ is the reward model
and $\rewardkernel{h}: (\Xcal \times \Acal)^{h} \to \Delta([-\Rmax, \Rmax])$. 

In this paper, we focus on learning policies that map from the most recent context to an action distribution, $\pi: \Xcal \to \Delta(\Acal)$. This is optimal when the domain is Markov and can often be more interpretable and more feasible to optimize given finite data in the offline setting. % **EB: we should discuss this a bit because this was a question from the reviewer
% The value of a policy is 
% $ v^{\pi} = \EE_{\transitiondescriptor, \rewarddescriptor} \left[ \sum_{h=1}^H r_h | a_{1:H} \sim \pi \right]$.
In offline policy optimization settings, we have a dataset with n trajectories collected by a fixed \emph{behavior} policy $\mu: \Xcal \to \Delta(\Acal)$, 
% $$
%     \left\{ \left\{ x_h^{(i)}, a_h^{(i)}, r_h^{(i)} \right\}_{h=1}^{H} | x_h^{(i)} \sim \transitionkernel{h}(x_{1:h-1}, a_{1:h-1}), a_h^{(i)} \sim \mu( x_h^{(i)}), r_h^{(i)} \sim \rewardkernel{h}(x_{1:h}^{(i)}, a_{1:h}^{(i)}) \right\}_{i=1}^{n},
% $$ 
and we aim to find a policy $\pi$ in a policy class $\Pi$ with the highest value.
% For the ease of notation, we use $\history_h$ to denote the history $(x_{1:h}, a_{1:h})$ and use $\history_h^{(i)}$ respectively. 
%This is related to the offline contextual bandit $(H=1)$ and offline reinforcement learning (RL) problems $(H > 1)$. 
% *EB: I don't think we need this overview here
%There are mainly three types of approaches. The (MDP) model-based approaches assume the transitions and reward kernels to be Markov and realizable by some function class. It learns approximate models then solves the policy by planning on the approximate models. The value-based approach assumes that there exists a Markov $Q$ function class with a low Bellman projection error such that the true $Q$ functions are realizable. % The core of $Q$ learning algorithm often involves a fitted iteration on the Bellman operator and an approximate/greedy policy to maximize the $Q$ function. 

Policy gradient and optimization approaches do not rely on a Markov assumption on the underlying domain, and have had some encouraging success in offline RL ~\citep{chen2019top}. 
%This work considers policy optimization approaches without modeling extra components such as value functions, transitions, and rewards. 
Often these methods leverage an importance sampling (IS) estimator in policy evaluation:  
$
    \is(\pi) = \frac{1}{n} \textstyle\sum_{i=1}^n \left( \textstyle\sum_{h=1}^H \ith{r_h} \right) \textstyle\prod_{h=1}^{H} \left( \frac{\pi(a^{(i)}_h|x^{(i)}_h)}{\mu(a^{(i)}_h|x^{(i)}_h)} \right).
$
The IS estimator is an unbiased and consistent estimate of the value under the following two assumptions: 
\begin{assumption}[Overlap]
For any $\pi \in \Pi$, and any $x \in \Xcal$, $a \in \Acal$, $\frac{\pi(a|s)}{\mu(a|s)} < \infty$.
\end{assumption}
\begin{assumption}[No Confounding / Sequential ignorability]
For any policy $\pi \in \Pi$ and $\mu$, conditioning on the current context $x_{h}$, the sampled action $a_h$ is independent of the outcome $r_{h:H}$ and $x_{h+1:H}$.
\end{assumption}

%The overlap assumption guarantees the importance weights always exist for any $\pi \in \Pi$. The no confounding (sequential ignorability) assumption \citep{namkoong2020off} assumes the action distribution under any policy is Markov. 
%Under these assumptions IS is an unbiased estimator, but still often suffers from high variance. 
IS often suffers from high variance, which has prompted work into extensions such as doubly robust methods~\citep{jiang2016doubly,thomas2016data} and/or methods that balance variance and bias.
% \begin{align}
%     \wis(\pi)  = \textstyle\sum_{i=1}^{n} \ith{r} \textstyle\frac{\ith{W}}{
%     \Wsum}, \quad \tis(\pi)  = \textstyle\frac{1}{n} \textstyle\sum_{i=1}^{n} \ith{r} \max \{ \ith{W}, M \},
% \end{align}
%With finite number of sample, these estimators is often more stable and enjoy smaller mean-square error empirically than the vanilla IS. We can also combine these two tricks to further control the variance in the importance weights, which is
Truncating the weights and using self-normalization has been shown to be empirically beneficial both in bandit and RL settings \citep{swaminathan2015self,futoma2020popcorn}: we refer to this as Self-Normalized Truncated IS (SNTIS):
\begin{align}
\label{eqn:sntis}
\textstyle
    \wtis(\pi) := \frac{\sum_{i=1}^n \left( \sum_{h=1}^H \ith{r_h} \right) \min\left\{  \textstyle \prod_{h=1}^{H} \ith{W_h}, M \right\}}{ \sum_{i=1}^n \min\left\{  \textstyle \prod_{h=1}^{H} \ith{W_h}, M \right\} },
\end{align}
where 
\begin{equation}
\label{eqn:w}
\ith{W_h} := \frac{\pi(a^{(i)}_h|x^{(i)}_h)}{\mu(a^{(i)}_h|x^{(i)}_h)}
\end{equation}
and $M$ is a constant that truncates the weights. For ease of notation in the rest of this paper, we define:\\
$\ith{W_{1:h}} := \prod_{h=1}^{h} \ith{W_h}$, $\ith{W} := \ith{W_{1:H}}$, $\Wsum = \sum_{i=1}^n \ith{W}$, and $\ith{r} = \sum_{h=1}^H \ith{r_h}$.

While this estimate can be used as a direct objective for off policy learning, it may still have a significant variance which is important when comparing across policies. Prior work in contextual bandits~\citep{swaminathan2015counterfactual,swaminathan2015self} included a variance penalization in the objective based on the  empirical Bernstein's inequality. 

Here we provide a simple extension to the multi-step setting to yield a target objective for offline policy learning: 
\begin{align}
\label{eq:crm}
    \textstyle\argmax_{\pi \in \Pi} \wtis(\pi) - \lambda \sqrt{\widehat{\var} \left( \wtis(\pi) \right) }.
\end{align}

%\begin{align}
%\label{eq:crm}
%    \textstyle\argmax_{\pi \in \Pi} \wtis(\pi) - \lambda \sqrt{\widehat{\var} \left( \wtis(\pi) \right) }.
%\end{align}

% ** can get into this later
% ** cite popcorn or the other paper 

\section{Related Work}
\label{sec:prior_work}
There is increasing interest in multi-armed bandits and offline RL to avoid overly optimistic estimates of policies computed from finite datasets that can cause suboptimal policy learning. In this paper, we will show a particular unaddressed issue with IS methods avoiding initial states that lead to poor outcomes. In contrast, prior work has shown how to use self-normalized IS (also known as weighted IS) to address over maximizing  bandit rewards~\citep{swaminathan2015self}.
%is  equivariant to any constant shift in rewards, and will still suffer from the context avoidance issues we describe above. 
Counterfactual risk minimization~\citep{swaminathan2015counterfactual,joachims2018deep} uses variance regularization based on the empirical Bernstein's inequality for bandit problems.  
However, this penalization is at the policy level. Both self-normalized IS (SNIS) and variance penalization do not directly solve the problem with avoiding contexts with low reward. In Figure~\ref{fig:overfitting_example_result} in Appendix~\ref{appendix:example}, we show the counterfactual risk minimization regularization with or without self-normalization requires a large dataset to perform well. 
%Later work \citep{joachims2018deep} extended norm-POEM to stochastic gradient descent settings for large-scale training. 
Recent work \citep{brandfonbrener2020offline} discussed a similar overfitting issue as we describe and compared the performance of offline policy optimization and model/value-based method on such issue in the bandit setting. Those authors primarily focus on the negative result of the policy optimization approach and the advantage of the model/value-based method, whereas our approach suggests a method for addressing this issue in policy optimization and focuses on the RL setting. Doubly robust estimators \citep{dudik2011doubly,jiang2016doubly,thomas2016data,kallus2019double,kallus2019efficiently} have multiple benefits but, as long as the learned $Q$-function is imperfect, the issue of avoiding low performing contexts can still remain as the methods may overfit to the high/positive residual $r - Q$. Pessimism under uncertainty approaches are promising~\citep{kidambi2020morel,yu2020mopo} but have so far only been developed for Markov settings and are not robust to model class misspecification, unlike IS-based policy optimization.

Another line of offline batch policy optimization constrains the policy search space to be close to the behavior policy, or requires the action taken to have some minimum probability under the behavior policy~\citep{kumar2019stabilizing,buckman2020importance,sachdeva2020off,fujimoto2019off,futoma2020popcorn,DBLP:conf/uai/LiuSAB19,liu2020provably}. This work has focused on algorithms and analysis for the Markov setting with additional model realizability and/or closure  assumptions that are hard to verify. As we will discuss and empirically validate later, such constraints on the expected or observed empirical behavior policy are not yet sufficient, at least in large state spaces. 

Our work can be viewed as following in the recent line of work on pessimism under uncertainty\cite{liu2020provably,yu2020mopo,kidambi2020morel,buckman2020importance}, but adapted to provide policy search based offline learning method that does not require the Markov assumption or model realizability, and achieves strong performance given a finite dataset.

% The setting does not satisfy the Markov assumption and so the value function or MDP model based approaches also fail in this setting. 
% This overfitting problem is related to the \emph{propensity overfitting} issue discussed in previous literature in contextual bandit settings. Over-weighing the weights for high-rewards contexts, as well as over-avoiding the weights for low-rewards contexts are the two sides of the same propensity (importance weights) overfitting issue: fitting the weights of context instead of the weights of actions. The norm-POEM \citep{swaminathan2015self} method use self-normalized IS estimator to avoid over-weighing high reward boundlessly. The self-normalization of weights cannot prevent over-avoiding the low-rewards contexts, which eventually also result in over-weighing high reward in a self-normalized estimator. CRM's variance penalty \citep{swaminathan2015counterfactual} method relieves the overfitting problem by the variance penalization. It combats both the unavoidable variance from reward/transition dynamics and the variance from importance weights, thus might fail to avoid the overfitting in importance weights when the variance from the environment is non-uniform. 

\section{State Propensity overfitting}
\label{sec:overfitting}
We identify an important potential issue with using IS estimators during offline policy optimization, as we illustrate in contextual bandits when using the SNIS estimator.

Let 
$v^\pi(x) = \mathbb{E}[r|x, a \sim \pi]$, $\hat{p}(x)$  be the empirical probability mass/density over the contexts $x$ in the dataset, and $W(x) = \sum_{i: x^{(i)} = x} \frac{W^{(i)}}{W}$ where $W$ is the importance weight (Eq.~\ref{eqn:w}). We now decompose the importance weighted off-policy estimator into three parts. 
\begin{align}
    \hat{v} &= \underbrace{\mathbb{E}_{\hat{p}}[v^{\pi}(x)]}_{\text{empirical }v} + \underbrace{\sum_{x \in \Xcal} (\hat{p}(x)-W(x)) v^{\pi}(x)}_{\text{difference in context weights}}\label{eqn:decomp1}\\
    &+ \underbrace{\sum_{x \in \Xcal} W(x) \left( \sum_{x^{(i)}=x} \frac{W^{(i)}}{W(x)W} r^{(i)} - v^{\pi}(x) \right)}_{\text{weighted IS error in each context}}
    \label{eqn:decomp}
\end{align}
The first term is a supervised empirical value estimate whose only error is due to the error in the empirical context distribution sampled in the dataset versus the true context distribution. The second term captures the error caused by the difference between context distribution introduced by weights and empirical context distribution in the dataset. The third term computes the difference between the weighted IS estimate of the value of the policy in a specific context $x$ versus its true value $v^{\pi}(x)$, and then sums this over all contexts.

The second term is of particular interest, because it highlights how the IS estimator of a policy may effectively shift the relative weight on the context space. In the bandit setting (and in the initial starting state distribution for RL), such shifting should not be allowed: the policy may control what actions are taken, but cannot change the initial context distribution. We now show how an algorithm maximizing the importance weighted off-policy estimate can exploit this structure and yield overly inflated estimates (Eq.~\ref{eqn:decomp}). 

\begin{example}
Consider a contextual bandit problem with $|\Xcal|$ contexts and $|\Acal|$ actions in each context. For half the contexts $S_p$, the reward is $1$ for one action and zero for others. For the other half of the contexts, $S_n$, the reward is -1 for half the actions, and -5 for the rest. The true distribution over contexts is uniform. The optimal policy would have an expected reward of 0 over the state space. The behavior dataset is drawn from a uniform distribution over contexts and actions.  When the sample size $|\Dcal| < |\Acal|$, we assume there is only one observed positive reward in the dataset. A policy $\pi_o$ that maximizes Eq.~\ref{eqn:decomp1}-\ref{eqn:decomp}  
will select actions that are not present in the dataset for all contexts whose observed actions lead to only zero or negative rewards: let $\mathcal{A}_{s_i} = \{a_i :  r(s_i,a_i) \leq 0 \}$ then $\pi(s_i) = a_j$ where $a_j \notin \mathcal{A}_{s_i}$. This will yield $W(x)=0$ on all contexts except for the contexts with observed positive rewards. The resulting IS/SNIS estimator of the value of $\pi_o$ is $1$, which is much higher than the optimal value 0. In addition, such a $\pi_o$ is likely to be worse  than the optimal policy for any context where  $r(s_i,a_i)=-1$, since that is the optimal reward possible for such states $s_i$, and by $\pi$ selecting an unobserved action $a_j$ in that state ($\pi(s_i)=a_j$), the policy $\pi$  may select an action with worse true reward,  $r(s_i,a_j)=-5$.
\end{example}

%This is a unique overfitting phenomena in the  counterfactual (or called off-policy) learning settings since in supervised learning it is not possible for a  hypothesis $\pi$ to manipulate the weights over the input contexts.

While this issue can arise in contextual bandits~\citep{swaminathan2015self}, it is even easier for this to occur in sequential RL (Examples \ref{example:rl} and \ref{example:rl_model_fail} in Appendix~\ref{appendix:example}). Intuitively, the issue arises because when estimating the value of a new decision policy, it is acceptable to choose a policy that re-distributes the \emph{weights of actions within an initial context} but not that re-distributes the \emph{weights across initial contexts}, since it is not a function of the actions selected. It is well known that in importance sampling, the expected ratio of the weights should be 1: $\EE_{y \sim \mu} [\pi(y)/\mu(y)] = 1$. In contextual policies, we expect that for each initial context $x_0$, the expected weights should also be 1: $\EE_{a \sim \mu(a|x_0)} \left[\pi(a|x_0)/\mu(a|x_0)|x_0\right] = 1$. 
However, optimizing for a standard importance sampling objective (such as Eq.~\ref{eqn:decomp}) does not involve constraints that the empirical expectation of weights given an initial context $\empE[\ith{W_h}|\ith{x_h}]$ (or the weights of $n$-step given initial context $\empE[\ith{W}|\ith{x_1}]$ ) is still close to one. 
%In order to optimize the future return, the optimization algorithm can optimize the weights of initial contexts in the dataset, instead of optimizing the weights on different actions/ action sequences in a given initial context. 

Such \textit{propensity overfitting} may seem surprising given that under mild assumptions, which are satisfied here by Assumptions 1 and 2, IS provides an unbiased estimate of a policy's value. Our observations do not contradict this fact: while IS will still provide an unbiased estimate given a policy, policy optimization can exploit the finite sample error and lack of data coverage.  

% how well their context-action distributions are represented in the available data, and policy optimization can exploit this. 

% discuss why previous work does not work

\begin{figure*}[ht]%{R}{0.8\textwidth}
    \centering
    \includegraphics[width=\textwidth]{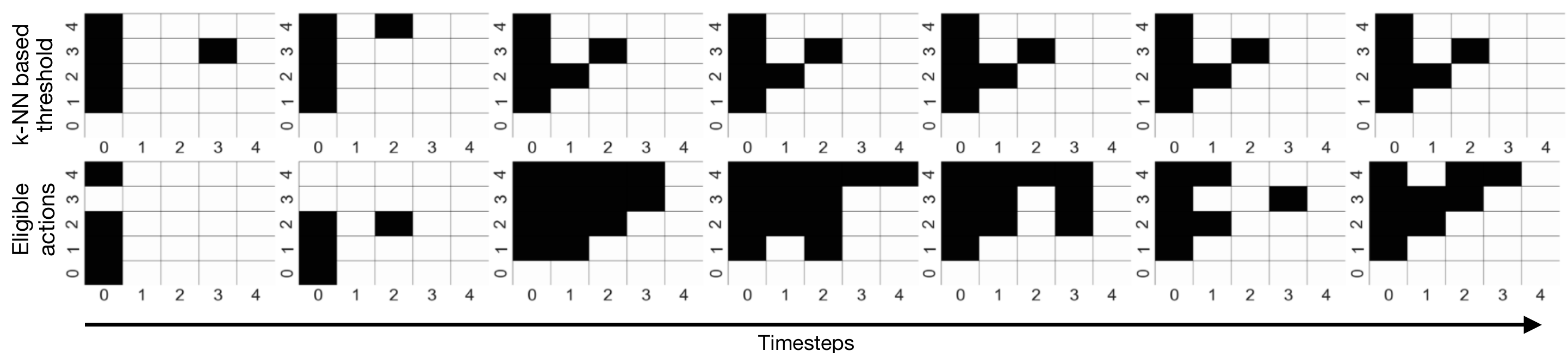}
    \caption{Different policy class constraints in MIMIC sepsis data. The top shows in black which of the 25 actions pass a constraint on having minimal probability over an estimated behavior policy, 
    $\mu_{KNN}(a|s) > b$, for a sequence of patient states. The bottom shows POELA's eligible actions for the same states. This illustrates that even constraints given the empirical estimated behavior policy can be very different than POELA's eligible actions. Note sometimes POELA is more conservative, and other times  the estimated behavior policy constraint is more conservative.}
    \label{fig:sepsis_traj_action_eligibility}
\end{figure*}

%\label{subsec:alt}
One might hope that existing methods are sufficient to address this challenge. Here we expand on the discussion in the related work to suggest why this is not the case. First, splitting the data into training and selection sets (e.g.,~\citet{thomas2015high,komorowski2018artificial}) is generally insufficient. If  some of the performance gains come from systematically avoiding actions taken in initial states with low performance, then it is likely that a similar performance benefit can also arise in the validation set if IS-based estimators are used both in policy selection and later estimation. We will observe experimentally this is true even when the estimator or objective involves a variance penalization (Eq.~\ref{eq:crm}). For example, this may occur when only a small set of initial states are avoided, in a way that only mildly impacts the variance and effective sample size, yet results in substantially overly optimistic estimates.

It is also insufficient to shift all rewards to be non-negative. While this voids the benefit of avoiding states if standard IS is used, popular lower-variance IS off-policy estimators like weighted IS are equivariant to any constant shift in rewards. Similarly, doubly robust estimators~\citep{jiang2016doubly,thomas2016data} frequently center rewards around estimates of the reward/value outcomes. 

%In both cases they can still suffer from the state context avoidance issues described above. 
\textit{Is constraining to the behavior policy sufficient?} Perhaps the most compelling idea is whether constraining the policy class to actions with some minimal probability under the behavior policy\footnote{A minimal requirement for consistent estimation of a target policy using IS is that there is overlap between the behavior policy and target, which we also assume.}. First note that constraining to the true behavior policy (e.g.,~\citet{fujimoto2019off,sachdeva2020off}) can still cause the propensity overfitting problems we described when the dataset is insufficient to cover all non-zero behavior probability actions in all states. In addition, sometimes the behavior policy is itself unknown. Estimating the behavior policy from data and using this in both the IS-based objective and overlap constraints may be more practical given datasets of limited size and when the state space is large. Indeed, semi-parametric theory and past related work in bandits~\citep{narita2019efficient} and RL~\citep{hanna2021importance} have suggested that even if the behavior policy is known, leveraging the estimated behavior policy can yield more accurate offline policy estimates. It is natural to assume such benefits might also translate to improvements for constrained policy learning. 

While promising in principle, this approach may be challenging in large state space environments. First, in such settings the maximum number of observed actions in any particular state is almost always one. Assuming there is good reasons to believe that the behavior policy is not actually deterministic, it is necessary to use some function approximation method to estimate the behavior policy~\citep{hanna2021importance}, which may be a deep neural network or non-parametric methods like k-nearest neighbors~\citep{raghu2018behaviour}. Unfortunately, as we will demonstrate in our experiments, we find that such approximators may sometimes be sufficient to accurately predict the behavior policy for a given state, but do not seem to be as beneficial when used to constrain the targeted policy class. Such estimates may overestimate (or underestimate) the probability of taking alternate actions in some states, and therefore enable both context avoidance or be too conservative in their policies. Figure~\ref{fig:sepsis_traj_action_eligibility} illustrates that our method can provide quite different constraints on the policy class than using constraints on the estimated empirical behavior policy, shown for a patient in the MIMIC sepsis dataset.

\section{Policy Optimization with ELigible Actions}
\label{sec:algorithm}
We have described that in IS-based estimators, because all weight can be placed on unobserved actions for certain initial contexts, the empirical conditional expectation of weights given an initial context $\empE[W|x]$ can be zero for such contexts, instead of $1$. To address this, one possibility is to constrain  the conditional expectation of weights given a context to be $1$ or lower bounded. However doing so in   infinite/continuous context spaces is subtle:  each context likely only appears once in the dataset, and requiring $\empE[W|x]=1$ would be equivalent to only allowing a policy that exactly matches the observed logged actions.

We now propose a slight relaxation of the above proposal. Absence constraints, IS-estimator-based policy learning can place a large weight on unobserved actions in the dataset, for which our reward uncertainty is high. The recent line of pessimism under uncertainty for model and value based MDP offline RL (e.g.,~\cite{liu2020provably,yu2020mopo}) explicitly accounts for such statistical uncertainty through constraining or penalizing actions and/or states and actions for which there are limited observed data. Our approach is similar but designed to address these issues in settings that may not be Markov. 

Specifically, for policy learning, we create local constraints on the eligible policy class. For each context $x$, dataset $\Dcal$ and a given threshold $\threshold$, the \emph{eligible action} set $\Aset(x;\Dcal,\threshold)$ is:
\begin{equation*}
    \Aset_h(x;\Dcal,\threshold) \!=\! \{ a_h: \exists (x_h,a_h) \!\in \!\Dcal \, s.t. \, \dist(x,x_h) \!\le\! \threshold \}.
    \end{equation*}
This allows only the action that was taken for a given context, or actions taken in contexts within a given distance $\delta$ of the observed contexts. Note that in large or continuous action spaces, the resulting allowed actions per context may be quite different than policy search methods  that places thresholds using the behavior policy probability~\citep{futoma2020popcorn} as we will observe empirically (cf. Figure~\ref{fig:sepsis_traj_action_eligibility}). Intuitively, our eligible action constraint constraints the  policy class to \textit{observed} actions taken in the current or nearby\footnote{In this work, we use the Euclidean distance on the state space, but for very high dimensional state spaces, it would likely be beneficial to compute distances leveraging  representation learning RL  work (e.g., \cite{zhang2020learning}). We also considered other approaches to defining a “neighborhood” of a given context: k-nearest neighbors. The problem was that this could include samples that were very far away in context space, and so it could appear like there was local similarity and coverage of the policy, but in reality no nearby states had taken a similar action.} states, but behavior policy thresholds can allow actions that could be taken at that state, even if no such action was taken there, nor at any nearby state. We will shortly prove that our eligible action set is sufficient to lower bound the empirical conditional expectation of weights per observed context. This ameliorates the propensity overfitting problem, and we will shortly show that empirically this can yield significant benefits. 

%The above results highlight that introducing a constraint on the policy class to have local overlap with actions taken in the dataset, is sufficient to ensure that the weights on the contexts are lower bounded. This will help address the overfitting issue highlighted in the prior sections. 

Policy learning can  be done by finding the best policy that satisfies the eligible actions constraints given the  dataset:
\begin{align}
\label{eq:poela_objective}
    \textstyle\argmax_{\pi \in \Pi} & \!J(\pi; \Dcal) \, \nonumber \\ s.t. & \, \forall i,h \, \textstyle\sum_{a \in \Aset_h(\ith{x_h};\Dcal,\threshold)} \!\pi(a|\ith{x_h}) \!= \!1.
\end{align}
$J(\pi; \Dcal)$ can be any objective function such as $\is$ or $\wtis$.% (Equation \ref{eqn:sntis}) or  the counterfactual risk minimization regularization (Equation \ref{eq:crm}).
% ----
% We assume that $J(\pi; \Dcal)$ does involve importance sampling as part of its objective: when the objective is a value-based or model-based MDP value function, existing methods for pessimism under uncertainty can be employed. % **EB: any benefit to this there?
\begin{algorithm}[tb]
  \caption{Policy Optimization with ELigible Actions (\ispg).}
  \label{alg:is_policy_optimization}
    \begin{algorithmic}[1]
  \State {\bfseries Input:} $\Dcal$, $\Pi_\theta$, sphere radius $\threshold$, IS truncation $M$, CRM coefficient $\lambda$, learning rate $\alpha$
  \State {\bfseries Output:} $\widehat{\pi}_{\theta}$
  \State Initialize $\theta_0$
  \For{$t=0, 1$ {\bfseries until convergence}}
  \State \vspace{-8pt}
    \begin{align*}
    \widehat{\pi}_{\theta_t}(a|x) := \frac{\ind\{ a\in \Aset_h(x;\Dcal, \delta) \} }{\sum_{a}  \ind\{ a\in \Aset_h(x;\Dcal, \delta) \} \pi_{\theta_t}(a|x)} \pi_{\theta_t}(a|x)
    \end{align*} \label{alg:action_eligibility}
  \vspace{-12pt}
  \State \vspace{-8pt}
  \begin{align*}
      \hspace{-8pt}\theta_{t+1} \leftarrow \theta_{t} + \alpha \nabla_{\theta} \left(  \wtis(\widehat{\pi}_{\theta_t}) - \lambda \sqrt{\widehat{\var} \left( \wtis(\widehat{\pi}_{\theta_t}) \right)  } \right)
  \end{align*}
  \EndFor
\end{algorithmic}
\end{algorithm}
We now present our \ispg (Policy Optimization with ELigible Actions) algorithm (Algorithm~\ref{alg:is_policy_optimization}) that implements the learning objective in Equation \ref{eq:poela_objective}. We use the counterfactual risk minimization objective function (Eq. \ref{eq:crm}) as the $J(\pi;\Dcal)$, where the estimator $\widehat{\var}(\wtis)$ is constructed using the Normal approximation in \citep[Equation 9.9]{owen2013monte}:
$ \widehat{\var}(\wtis) = \frac{\sum_{i=1}^n \left(\ith{r} - \wtis \right)^2 (\min\{\ith{W},M \})^2 }{\left( \sum_{i=1}^n  \min\{\ith{W},M \}\right)^2}. $

After each gradient step, we enforce the policy to satisfy the eligible action constraints by re-normalizing the output probability on $\Aset_h(x;\Dcal, \delta)$ for $x \in \Dcal$. The eligible action set for each training sample is static and can be stored to reduce computational cost. In the experiments, we use the Euclidean distance over nearby states at any time index.%, and remove the constraints on the same time index $h$ eligible actions. % \ispg

\section{Analysis} 
\label{sec:overfitting_theory}
 
%Lower Bound of \texorpdfstring{$\mathbf{\empE[W|x]}$}{E[W|x]} and Eligible Actions}

In this section we formalize the benefit of the eligible action constraints, and also prove consistency guarantees on the resulting value estimates used in the objective, Equation~\ref{eq:poela_objective}. 

Eligible action constraints were introduced to help alleviate propensity overfitting, which can be characterized by the expected empirical sum of propensity weights for a particular context $x$ being much lower than 1, or even 
$\mathbf{\empE[W|x]} = 0$. We now show that any policy that only selects actions in eligible action sets will ensure that the empirical sum of weights in any hypersphere of any context can be lower bounded, as desired. As mentioned before, for very large state spaces, where only a single action is observed for each observed state in the dataset, the only way to ensure that $\mathbf{\empE[W|x]} = 1$ is to reduce the policy to the observed actions. Intuitively, the guarantee we provide here is reasonable when local smoothness is present and a soft form of state aggregation is tenable, ensuring that the empirical expected sum of weights over nearby states is lower bounded. 

To do so, we first introduce an assumption about the target policy's smoothness in the context space.
\begin{assumption}[$L$-Lipschitz policy]
$\forall \pi \in \Pi$, $ \| \pi(a|x) - \pi(a|x') \| \le L\dist(x,x') $.
\end{assumption}
That is, nearby contexts have similar actions~\citep{berkenkamp2017safe,wang2019generalization} which ensures that we have some minimal weight support in a small neighborhood. If different policies have different smoothness, the Lipschitz constant can be taken to be the max over the policy set. Under this assumption and the former assumptions, we can show the following. Proofs are provided in Appendix~\ref{appendix:theory}:
\begin{theorem}
\label{thm:weights_lowerbound} $\forall \ith{x_h}$, $\Bcal(\ith{x_h},\threshold) := \{x: \dist(x,\ith{x_h}) \le \threshold\}$,
$\sum_{ x_{h}^{(j)} \in \Bcal(\ith{x_h},\threshold) } W_h^{(j)} \ge 1 - \threshold  L |\Acal| $.
\end{theorem}

Given the likelihood ratio is lower bounded, we can further show that the self-normalized truncated weights are also lower bounded in the one-step settings.
\begin{corollary}
\label{cor:onestep_weights_lowerbound}
For $H=1$,
$
    \sum_{x_{1}^{(j)} \in \Bcal(\ith{x_1},\threshold)} \frac{\max\{\ith{W},M \}}{
     \sum_{i=1}^n \max\{ \ith{W},M \} } \ge \frac{1-\threshold L |\Acal|}{nM}
$ for $M > 1$.
\end{corollary}
In the $n$-step sequential setting, it is necessary to have the 1-step weights be  greater than zero in order to have $n$-step weights greater than zero. 
% $\EE [W_{1:h}^{(i)} | x_{h}^{(i)} \in \Bcal(x,\threshold) ] = \EE [\ith{W_{1:h-1}} | x_{h}^{(i)} \in \Bcal(x,\threshold) ] \EE [ \ith{W_h} | x_{h}^{(i)} \in \Bcal(x,\threshold) ]  $.  It is true for the empirical mean as well.
\begin{proposition} 
\label{prop:nstep_necessity}
For any $x$, $\delta$, $\EE [W_{1:h}^{(i)} | x_{h}^{(i)} \in \Bcal(x,\threshold) ] = \EE [\ith{W_{1:h-1}} | x_{h}^{(i)} \in \Bcal(x,\threshold) ] \EE [ \ith{W_h} | x_{h}^{(i)} \in \Bcal(x,\threshold) ] $. $\empE [W_{1:h}^{(i)} | x_{h}^{(i)} \in \Bcal(x,\threshold) ] = \empE [\ith{W_{1:h-1}} | x_{h}^{(i)} \in \Bcal(x,\threshold) ] \empE [ \ith{W_h} | x_{h}^{(i)} \in \Bcal(x,\threshold) ]. $
\end{proposition}

%If the policy overfits the weights on an h-step context such that $\empE [ \ith{W_h} | x_{h}^{(i)} \in \Bcal(x,\threshold) ] = 0 $, the $n$-step weights will also be zero even if the roll-in probability under $\pi$ is non-zero. 

The action eligibility local constraints provide a conservative pessimism-like constraint on the policy class, ensuring that the policy does not take actions which have not been tried in nearby states (and therefore for which the potential outcomes are unknown). As we will see shortly, this will yield more stable and beneficial performance in our simulations. An additional desirable property is that asymptotically, POELA relaxes to unconstrained policy learning. 
\begin{theorem}
\label{thm:consistV}
(Contains all overlapping policies). For a fixed $\threshold$, for any $x$, $\Aset_h(x;\Dcal,\threshold) \to \{a: \mu(a|x) > 0\} $ as $n \to \infty$ with probability 1. Therefore asymptotically the policy class will contain  all $\pi$ satisfying the overlap assumption.
\end{theorem}
\begin{theorem}
\label{thm:consistency} % For a fixed $\threshold$, as $n \to \infty$, the solution to Equation \ref{eq:poela_objective} is the same as $\hat{\pi}_{\Dcal,J} = \argmax_{\pi} J(\pi,\Dcal)$, over the policy class $\Pi$ such that $\forall \pi \in \Pi$, $\pi(a|s) > 0$ only if $\mu(a|x) > 0$. ii) 
Let $J(\pi,\Dcal)$ be the objective \footnote{Other consistent estimators $J$ can also be shown to satisfy this property, such as IS and self-normalized IS.} in Equation \ref{eq:crm}, the truncation threshold $M$ as a function satisfies $M \to \infty$ and $M/n \to 0$ as $n \to \infty$, and $|\Pi|<\infty$, then $v^{\hat{\pi}_{\Dcal,J}} \to \max_{\pi \in \Pi} v^{\pi}$ in probability.
\end{theorem}

%A natural question is whether these action eligibility local constraints limits the expressivity of the policy class $\Pi$. We show that asymptotically the expressivity is the same, and the maximizer in Equation \ref{eq:poela_objective} will converge to the optimal policy in the policy class if $J(\pi,\Dcal)$ is consistent.%\footnote{Note that importance sampling and weighted importance sampling are both consistent estimators, as is self-normalized truncated IS if the truncation $M$ is set as an increasing function of the dataset size.}
%\begin{theorem}
%\label{thm:consistency}
%i) Fixed $\threshold$, for any $x$, $\Aset_h(x;\Dcal,\threshold) \to \{a: \mu(a|x) > 0\} $ as $n \to \infty$ with probability 1. Thus the solution to Equation \ref{eq:poela_objective} is the same as $\hat{\pi}_{\Dcal,J} = \argmax_{\pi} J(\pi,\Dcal)$. ii) If $J(\pi,\Dcal)$ is the objective in Equation \ref{eq:crm}, the truncation threshold $M$ as a function satisfies $M \to \infty$ and $M/n \to 0$ as $n \to \infty$, and $|\Pi|<\infty$, then $v^{\hat{\pi}_{\Dcal,J}} \to \max_{\pi \in \Pi} v^{\pi}$ in probability.
%\end{theorem}

\section{Experiments}
\label{sec:experiment}

\begin{table*}[ht]
    \centering
    \begin{small}
    \tabcolsep=0.1cm % reduce table size by reducing column separation
    \begin{tabular}{c|c|ccccc|c}
        \toprule 
         & Algorithms & \ispg & \ispgknn & \ispgbaseline & BCQ & PQL & $9$-mon \\
         \midrule 
         Non-MDP & Test $v^\pi$ & $ 95.92 \pm 1.68 $ & $76.99 \pm 13.80$ & $77.32 \pm 14.55$ & $13.60 \pm 0.15 $ &  $19.64 \pm 5.71$ & $68.12 $ \\
         & $\wtis - v^\pi$ & $ -1.28 \pm 1.93$ & $16.07 \pm 13.55$ & $15.71 \pm 14.30$ &  $80.54 \pm 1.42 $ & $74.48 \pm 6.23$ & $-$ \\
        %  Drug dosing & $5.31 \pm 0.31 $ & $3.81 \pm 1.00$ & $\textbf{0.80} \pm 0.72$ & $2.56 \pm 0.56$ & $9$ \\
        \midrule
        MDP & Test $v^\pi$ & $89.53 \pm 1.32$ & $69.18 \pm 10.17$ & $63.27 \pm 13.27$ & $ 82.68\pm15.19 $ &  $99.98 \pm 0.38$ & $68.12$ \\
        & $\wtis - v^\pi$ & $ 5.12 \pm 2.01$ & $24.92 \pm 9.71$ & $30.82 \pm 12.59$ & $14.76 \pm 14.89  $ & $-2.65 \pm 1.76$ & $-$ \\
         \bottomrule
    \end{tabular}
    \end{small}
    \caption{LGG Tumor Growth Inhibition simulator. Test $v^\pi$ (1000 rollouts in the simulator) and $\wtis - v^\pi$ (amount of overfitting of the learned policy) with $\wtis$ on the validation set. %Drug dosing and treatment effect measured by MTD change is computed from the test rollouts. 
    Average across 5 runs with standard error reported.}
    \label{tab:tumor_result_main}
    % \vspace{-0.3cm}
\end{table*}

We now compare \ispg with several prior methods for offline RL. Perhaps the most relevant work in avoiding overfitting when using importance sampling is norm-POEM~\citep{swaminathan2015self}. %, that uses a  counterfactual risk minimization objective function for offline contextual bandits. 
For it to be suitable for sequential decision settings, we use a neural network policy class and refer to the resulting algorithm as~\ispgbaseline. A second baseline PO-$\mu$ constrains the policy class to only include policies which take actions with a sufficient probability under the behavior policy $\mu(a|s)$~\citep{futoma2020popcorn}. We also compare with recent deep value-based MDP methods in batch RL: BCQ \citep{fujimoto2019off} and PQL \citep{liu2020provably}. For all algorithms, we use a feed-forward neural network for the relevant policy and/or value function approximators. We report the test performance of the selected policy either through online Monte-Carlo estimation if a simulator is available, or using SNTIS estimates on a held out test set. Full details are provided in Appendix~\ref{appendix:experiment-details}.
 %including steps between checkpoints, hyper-parameters and their possible values, and the number of re-starts
% details: sample size, checkpoints, hyperparameters, random seeds

\subsection{Tumor Inhibition Simulator}
The Tumor Growth Inhibition (TGI) simulator \citep{ribba2012tumor} describes low-grade gliomas (LGG) growth kinetics in response to chemotherapy in a horizon of 30 steps (months), with a non-Markov context and a binary action of drug dosage~\citep{yauney2018reinforcement}. The reward is an immediate penalty proportional to the drug concentration, and a delayed reward of the decrease in mean tumor diameter. The behavior policy selects from a fixed dosing schedule of 9 months (the median duration from \citet{peyre2010prolonged}) with 70\% probability and else selects actions at random.

In this experiment, the behavior policy can only take values in $\{0.15, 0.85\}$. This means that constraining the policy class to have a minimal probability under $\mu(a|s)$, as in baseline \ispgknn, is only a non-trivial constraint for thresholds greater than $0.15$: this generates a single potential target policy, which is the deterministic fixed-dosage part of behavior policy. We include this as $9$-mon (short for 9 month dosing) in Table \ref{tab:tumor_result_main}. The training and validation sets both have 1000 episodes. We repeat the experiment 5 times with 5 different train and validation sets. Policy values are normalized between $0$ (uniform random) and $100$ (best policy from online RL). As shown in Table \ref{tab:tumor_result_main} (Non-MDP rows), \ispg achieves the highest test value as well as smaller variability compared with the baselines.

\textbf{Does \ispg reduce overfitting?} Examining the difference between $\wtis$ on the validation set and the online test value, we observe that most algorithms result in a policy whose value is a significant overestimate of its true performance (cf. Table\ref{tab:tumor_result_main}, $\wtis - v^\pi$). In contrast, \ispg yields a policy whose value is much more accurately estimated and performs better. Experiments with final policies selected during training based on SNTIS estimates on the validation set suggest the same conclusion (cf. Table~\ref{tab:tumor_result_checkpoint} in Appendix~\ref{ap:tumor-checkpoint}).

%reflect the overfitting particularly due to the off-policyness and is different from the standard loss overfitting in supervised learning. 

\textbf{Performance comparison in a MDP environment.} We also repeat the experiment with an MDP modification of the simulator, including an immediate Markovian reward and additional features for a Markovian state space. Note that we expect BCQ and PQL to do very well: both are designed to avoid overfitting in offline MDP learning and in particular PQL uses a pessimism under uncertainty approach to penalize policies that put weight on state-action pairs with little support. Although \ispg makes no Markov assumptions, it ponly erforms on average slightly worse than the two conservative MDP methods but still outperforms BCQ. \ispg also substantially outperforms other policy classes.
%that do not rely on the Markov assumption.

% \textbf{Is the learned policy medically reasonable?} We also reported the amount of drug dosing (number of months with full dose) and the percentage of change in MTD. Compared with a fixed dosing of 9 month with MTD decrease $49.84\%$, our algorithm decrease MTD by $-38.04\%$ with only $5.49$ months of using drug. Other RL algorithm take less drug dosing but have a significant disadvantage in the MTD decrease.

\begin{table*}[ht]
    \centering
    \begin{small}
    % \vspace{-0.5cm}
    \begin{tabular}{c|ccccc|c}
        \toprule 
         Method & \ispg& \ispgmuhat & \ispgbaseline & BCQ  & PQL & Clinician \\
         \midrule 
        Test SNTIS & 92.32 (90.87) & 90.21 & 86.89 & 25.62 & 27.04 & 81.10 \\
        $95\%$ BCa UB & 95.83 (92.94) & 93.27 & 89.68 & 41.93 & 42.45 & 82.19 \\
        $95\%$ BCa LB & 90.91 (87.22) & 87.19 &83.50 & 7.93	& 13.43 & 79.80\\
        Test ESS & 437.03 (396.71) & 297.84 & 289.10 & 206.63 & 217.54 & 2995 \\
         \bottomrule
    \end{tabular}
    \caption{MIMIC III sepsis dataset. Test evaluation, $(0.05, 0.95)$ BCa bootstrap interval, and ESS. The value of \ispg without a CRM variance penalty is in parentheses.}
    \label{tab:sepsis_result_main}
    \end{small}
    % \vspace{-.2cm}
\end{table*}

\setlength{\tabcolsep}{4pt}{
\begin{table*}[h!]
    \begin{small}
        \hspace{-.9cm}
    \begin{minipage}{.5\linewidth}
    \centering
    \begin{tabular}{c|ccccc}
        \toprule 
         Method & \ispg& \ispgmuhat & \ispgbaseline & BCQ  & PQL \\
         \midrule 
        Test SNTIS & 86.42 (85.26) & 84.39 & 79.71 & 32.83 & 34.69  \\
        $95\%$ BCa UB & 91.68  (90.32) & 87.74 & 89.33 & 53.50 & 52.15  \\
        $95\%$ BCa LB & 79.71  (77.15) & 80.01 &65.01 & 11.87	& 17.60 \\
        Test ESS & 310.23 (287.39) &244.97  &224.92  & 207.12 & 223.03 \\
         \bottomrule
    \end{tabular}
    \caption{Idem except using behavior policy $\hat{\mu} = \text{BC}$.}
            \label{tab:sepsis_result_bc}
    \end{minipage}
            \hspace{+.9cm}
    \begin{minipage}{.5\linewidth}
    \centering
    \begin{tabular}{c|ccccc}
        \toprule 
         Method & \ispg& \ispgmuhat & \ispgbaseline & BCQ  & PQL  \\
         \midrule 
        Test SNTIS & 88.83 (88.31) & 87.97 & 85.33 & 33.21 & 41.66  \\
        $95\%$ BCa UB & 93.23  (94.04) & 91.17 & 89.44 & 63.17 & 57.99 \\
        $95\%$ BCa LB & 83.43  (80.02) &82.00  &78.43 & 12.23	& 14.76 \\
        Test ESS & 379.18  (265.36) &  220.74 &236.78  & 203.89 & 224.33  \\
         \bottomrule
    \end{tabular}
    \caption{Idem except using behavior policy $\hat{\mu} = \text{BCRNN}$.}
        \label{tab:sepsis_result_bcrnn}
    \end{minipage}
    \end{small}
    % \vspace{-.2cm}
\end{table*}}

\subsection{MIMIC III Sepsis ICU data }
\label{sec:sepsis}
Next, we apply our method in a real-world example of learning policies for sepsis treatment in medical intensive care units (ICU). We used an extracted cohort \citep{komorowski2018artificial} of patients fulfilling the sepsis-3 criteria from the MIMIC III data set \citep{johnson2016mimic} and obtained a dataset of 14971 patients, 44 context features, 25 actions and a 20 step maximum horizon. Full details are in the Appendix~\ref{ap:sepsis-details}. 
We hold out $20\%$ of data for validation and $20\%$ of data for the final test. Treatment logs do not include the probabilities of clinicians' actions. Instead, as suggested by prior work~\citep{raghu2018behaviour}, we estimate the probabilities of the behavior clinicians' policy by $k$-NN with $k=100$.  %As $\hat{\mu}(a|s)$ is no longer always positive, 
To ensure overlap, for all policy optimization algorithms we allow $\pi(a|s) > 0$ only if $\hat{\mu}(a|s) > 0$. Using SNTIS to evaluate the performance on a test set is appealing because it makes little assumptions on the underlying domain. But if only a few test behavior policy trajectories match a test policy, the resulting value estimate is likely unreliable. We measure the amount of overlap between the test set and a desired policy by the effective sample size (ESS) \citep{owen2013monte}. Only policies with an ESS of at least $200$ on the validation set are considered.\footnote{The variance penalty may not ensure that the ESS is large, because it is only a soft penalty rather than a constraint that ensures a minimum ESS.}%, e.g., if only 2 trajectories in a dataset match a desired target policy (ESS=2). When they have the same reward and weights, the variance penalty of self-normalized estimator will be 0.} 
Similar to prior work~\citep{thomas2015higha}, in addition to the SNTIS estimator on the test set, we also report a $95\%$ upper and lower bound from bias-corrected and accelerated (BCa) bootstrap. The clinician's column is the test dataset rewards and sample size.

Table~\ref{tab:sepsis_result_main} shows \ispg is the best on all metrics, achieving the highest evaluation on the test set, the highest upper and lower bounds, and the highest ESS. %To further confirm this, additional results with the procedure of picking the best policy during training based on the SNTIS estimates on the validation set are reported in Table~\ref{tab:sepsis_result_checkpoint} in Appendix~\ref{ap:sepsis-checkpoint}.
%\textbf{Is the variance regularization helpful?} 
We also show that \ispg's test performance without its variance penalty is worse than using it but is still higher than the  baseline algorithms. In Appendix~\ref{ap:sepsis-tradeoff}, we further detail the differences in the methods under the prism of ESS and performance. We also demonstrate in Appendix~\ref{ap:sepsis-action-viz} that \ispg takes actions which more closely match the clinicians' actions for patients with initially high logged SOFA scores (measuring organ failure) in the test dataset in comparison to other baselines, suggesting that propensity overfitting may be occurring more in other methods. Finally, Figure~\ref{fig:sepsis_traj_action_eligibility} illustrates different actions constraints considered in this paper, using a sample trajectory in the test set as an example, where the 25 actions are depicted in 5x5 grids.% We observe the eligible actions approach employed by \ispg allows broader coverage of the action space and more state-by-state diversity.

%While these results are promising, %we note that due to the size of the dataset, the evaluation metric can still have significant variance. 
%before deploying any offline reinforcement learning algorithm in a high stakes clinical setting, further investigation and collaborations with clinicians would be essential. To  further inform in this context, we have also set up for the MIMIC III sepsis dataset experiments with the procedure of picking best policies during the training based on the SNTIS estimates on the validation set. The results, reported in Table~\ref{tab:sepsis_result_checkpoint} in Appendix~\ref{ap:sepsis-checkpoint}, further confirm \ispg achieves strong results compared to other baselines. The context space is likely to be non-Markov in the logged features, and policy optimization methods consistently perform better than value-based MDP offline RL algorithms.

\subsection{Behavior policy Estimation}
We now explore further if constraining policies to be close to the empirical behavior policy may produce similar benefits, and whether this depends on the function approximator used. We consider two additional function approximators: (1) learning a deep neural network representation of the behavior policy using Behavior Cloning (BC), an imitation learning approach~\citep{pomerleau1991efficient} and (2) training a recurrent neural network behavior representation using BCRNN, a variant of BC with a RNN as the policy network. BCRNN can learn temporal dependencies, which can be helpful. More details are included in Appendix~\ref{ap:bc-details}.

Results in the MIMIC III sepsis dataset are shown in Tables~\ref{tab:sepsis_result_bc} and~\ref{tab:sepsis_result_bcrnn}. Results in the Tumor simulator are included in Tables~\ref{tab:tumor_result_bc} and~\ref{tab:tumor_result_bcrnn}. Overall, this behavior policy modelling modification impacts all methods, but \ispg still outperforms other baselines. %, taking advantage of its action selection constraint from observed data. 
We note that BCQ and PQL benefit from these alternate behavior policy approximators, while policy-based methods suffer from it in the Non-MDP setting.  Comparing the benefits of using BC versus BCRNN, BCRNN behavior policy approximators in the non-Markov settings generally helps, as expected. We report additional results where best policies are selected from checkpoints during training based on SNTIS estimates in Tables~\ref{tab:tumor_result_bc_checkpoint} and~\ref{tab:tumor_result_bcrnn_checkpoint} (tumor) and in Tables~\ref{tab:sepsis_result_bc_checkpoint} and~\ref{tab:sepsis_result_bcrnn_checkpoint} (sepsis). In this application-driven selection procedure, \ispg still yields  higher test values.

\subsection{Experiment with continuous state space}
In the next experiment, we use the OpenAI Gym environment~\citep{brockman2016openai} CartPole. We also apply our method to a non-Markov modification of the environment. More details about this experiment in Appendix~\ref{ap:more-exp-cartpole}. In these experiments, only policies with an ESS of at least $30$ on the validation set are considered. Because of space constraints, the full results are provided in Appendix~\ref{ap:more-exp-cartpole}. Tables~\ref{tab:cartpole_results} to~\ref{tab:cartpole_pomdp_results_checkpoint} show the results. We observe that in both MDP and Non-MDP settings, \ispg provides improved performance over other methods. We also provide some results on D4RL~\citep{fu2020d4rl} datasets in Appendix~\ref{ap:more-exp-d4rl}. These results show that relying on observed data to decide on action eligibility can be beneficial for learning from the relatively few number of trajectories collected by the behavior policy in a continuous state space.

% \section{Related Work}
% \label{sec:related_work}
% \input{related_work}

\section{Discussion  \& Conclusion}
\label{sec:discussion}
A natural question is whether POELA, in addition to its overall improved performance, reduces context avoidance/propensity overfitting in practice. In Appendix 
\ref{sec:is_low}, we find that POELA generally puts more weight on initial states with low observed outcomes than other IS policy optimization methods, suggesting that it addresses the motivating problem. POELA also does not seem highly sensitive to the threshold used in the eligible action constraint (cf. Appendix~\ref{sec:effect_delta}) although middle ranges are more effective.

An alternative to constrained optimization is a soft penalty based on the proportion of contexts that are avoided through selecting alternative actions. This idea was previously proposed for contextual bandits~\citep{sachdeva2020off}. This is challenging to approximate in the RL setting, where deficiency can occur at any steps in a trajectory: exploring this is an interesting area for future work. Another interesting direction is to adapt the solution found in~\citet{joachims2018deep} when using a minibatch biases the SNTIS estimate.

To conclude, we identify a new overfitting problem arising when using IS as part of an offline policy learning objective. To address this, we constrain the policy class to only consider logged actions taken by nearby states. This can be viewed as a pessimism constraint similar to the one used in MDP offline policy learning, but developed for a non-Markov, direct policy search setting. Our approach yields strong performance relative to state-of-the-art approaches in a tumor growth simulator, a real-world dataset on ICU sepsis treatment and in classic continuous control with few demonstrations. POELA may be particularly useful for many applied settings such as healthcare, education and customer interactions, which have a short/medium length decision horizon, but are unlikely to be Markov in the observed per-step variables. Leveraging constraints on an empirical behavior policy was not as helpful, but 
%For example, \citet{hanna2021importance} demonstrated that learning a behavior policy that maximized the likelihood of the logged data in a IS-based estimator did not yield the most accurate estimate. 
an interesting direction for future work is whether other ways of learning such behavior policy might yield additional benefits to our locally constrained approach.% Another interesting open question is extending eligible actions for continuous action spaces.

%Our empirical results on a medical tumor growth simulator and on a sepsis ICU treatment dataset, shows significant benefits 

%We study an overfitting problem related to the importance weights in offline policy optimization. We identify that the overfitting is caused by optimizing the importance weights such that empirical weights on each context are away from the expectation. To prevent that, we proposed an algorithm to constrain the choice of action to be close to the logged actions in the dataset. Thus we show that the weights are lower bounded in each context's neighborhood, and the context (with lower reward) in the dataset will not be eliminated by overfitting the weights. We provide an experiment on a medical tumor growth simulator and another real-world dataset of sepsis treatment in ICUs. In these two domains the proposed algorithm outperforms prior work to avoid similar overfitting issues, and the value-based batch RL baselines significantly. 

\begin{acknowledgements} 
Research reported in this paper was sponsored in part by NSF grant \#2112926 and the DEVCOM Army Research Laboratory under Cooperative Agreement W911NF-17-2-0196 (ARL IoBT CRA). The views and conclusions contained in this document are those of the authors and should not be interpreted as representing the official policies, either expressed or implied, of the Army Research Laboratory or the U.S. Government. The U.S. Government is authorized to reproduce and distribute reprints for Government purposes notwithstanding any copyright notation herein.

The first and last authors acknowledge the Simons Institute for the Theory of Computing as this research is initiated when the two authors were a visiting student and long-term participant, respectively, at the Theory of Reinforcement Learning program of the Simons Institute.
\end{acknowledgements}

\bibliography{liu_683}

\newpage
\appendix
% \section{Example of Model}
\onecolumn

We first briefly describe the structure of the Appendix here. In Appendix \ref{appendix:example} we add two more examples in the multi-step settings as supplementary to the example in Section \ref{sec:overfitting}. In Appendix \ref{appendix:theory} we provide the proofs of theorems in Section \ref{sec:overfitting_theory}. In Appendix \ref{appendix:experiment-details}, we include more experiment details. In Appendices~\ref{ap:more-exp-tumor},~\ref{ap:more-exp-sepsis},~\ref{ap:more-exp-cartpole} and~\ref{ap:more-exp-d4rl} we include more results in the considered domains including experiments with estimating the behavior policy with function approximation and experiments with an alternative policy selection procedure with best intermittent policy checkpoint and the D4RL dataset. In the real world dataset on ICU sepsis treatment, we also include in Appendix~\ref{ap:sepsis-ablation} an ablation study without ESS constraints for hyperparameter selection on the validation set and in Appendix~\ref{sec:effect_delta} an investigation of the effect of eligible action constraints $\delta$. In Appendix\ref{sec:is_low} we also investigate the the weight given by different methods to states with low observed outcomes, and we conduct experiments on the differences in the methods under the prism of ESS and performance in Appendix~\ref{ap:sepsis-tradeoff}. Finally, in Appendix~\ref{ap:sepsis-action-viz} we include visualizations of eligible actions for high/mid/low-SOFA patients in addition to a timestep-by-timestep visualization of the two action constraints considered in this paper (based on the eligible action set in \ispg and based on the probability under the behavior policy for other methods).

\section{Counter Examples in RL Settings}
\label{appendix:example}
\begin{figure*}[th]%{R}{0.8\textwidth}
%\vspace{-\intextsep}
%\hspace*{0.5cm}
    \begin{minipage}{0.33\textwidth}
    \centering
    \includegraphics[width=0.9\textwidth]{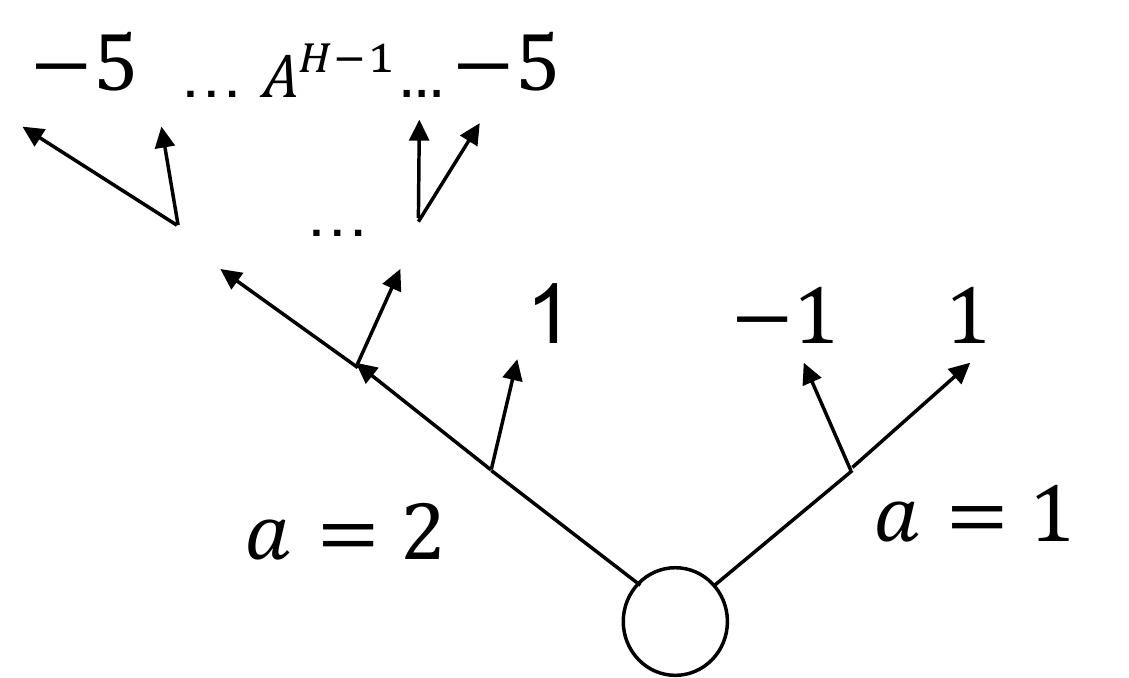}
    \subcaption{Example \ref{example:rl}.}
    \label{fig:overfitting_example}
    \end{minipage}
    \begin{minipage}{0.33\textwidth}
    \centering
    \includegraphics[width=0.9\textwidth]{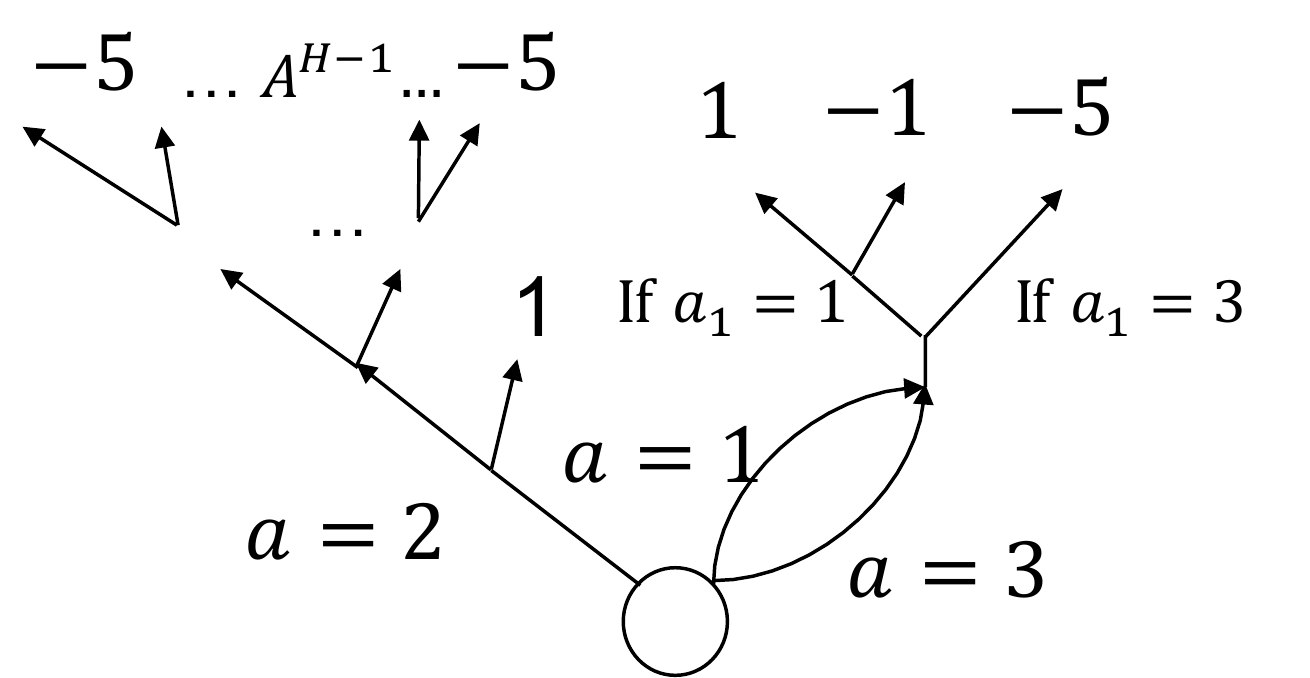}
    \subcaption{Example \ref{example:rl_model_fail}.}
    \label{fig:example_patient_model_fail}
    \end{minipage}
    \begin{minipage}{0.33\textwidth}
    \centering
    \includegraphics[width=0.9\textwidth]{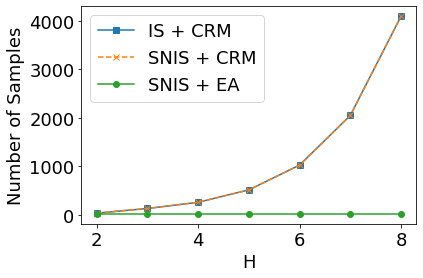}
    \subcaption{Number of samples to solve Example 2/3 for $A=2$.}
    \label{fig:overfitting_example_result}
    \end{minipage}
\end{figure*}

In the main text, we gave an example about the overfitting issue in contextual bandits with large state and action space in small datasets. Here we show that it is even easier for this to occur in sequential reinforcement learning settings, even when only 2 actions are available in the next two examples with or without state aliasing.
\begin{example}
\label{example:rl}
Consider a sequential treatment problem as shown in Figure \ref{fig:overfitting_example}. There are two actions available in each state. From the first state,
action $a_1$ has a 50\% chance of leading to an immediate terminal positive reward $r=1$ and a 50\% chance of leading to an immediate terminal negative reward $r=-1$. From the first state, action $a_2$ also has 50\% chance of leading to an immediate terminal positive reward $r=1$. For the other 50\% of states, action $a_2$ results in transitions to additional states, which are followed by additional actions, for another $H-1$ steps; however, all transitions eventually end in a large negative outcome (e.g., $r=-5$). For example, one could consider a risky surgical procedure that results in many subsequent additional operations and but is ultimately typically unsuccessful. Assume the behavior policy is uniform over each action, yielding  $\mu(a=0|x_1) = \mu(a=1|x_1)$ = 0.5 and a probability of each action sequence following $a_2$ of $\frac{1}{|A|^{H-1}}$. With even minimal data the value of $\pi(x_0)=a_1$ will be accurately estimated as 0. However, when  $H$ is large relative to a function of the dataset size, there always exists a action sequence after an initial selection of $a_2$ that is not observed in the dataset. This means that a policy $\pi_2$ that starts with $\pi(x_0)=a_2$ and then selects an unobserved action sequence will essentially put 0 weight on the resulting contexts that incur $r=-5$ outcomes, even though such outcomes will occur 50\% of the time after taking action $a_2$.  In this case, the value of $\pi_2$  will be overestimated significantly by IS or self-normalized IS. Thus the offline policy optimization will prefer taking action 2 at the first step as a result of overfitting even though the true value of first taking $a_2$ is $-1.5$ and the optimal policy value is $0$, obtained by taking action $a_1$. 
%We need to make the decision between two actions in the first step. Both actions have a positive treatment effect on half of the patients leading to $r=1$. Action 1 has a side effect ($r=-1$) on the other half of the patients. Action 2 has a stronger side effect ($r=-5$) on the other half of the patients, however, not immediately observed. These patients will be treated with $H-1$ more steps with $A$ actions, leading to $A^{H-1}$ different action sequences. In this example, data are draw from an uniform random policy $\mu(a=0|x_1) = \mu(a=1|x_1)$ = 0.5, and the probability of each action sequence following $a=2$ is $\frac{1}{A^{H-1}}$. When $A$ and $H$ are large, there always exists a sequence that is not observed in the dataset. In this case, the policy taking action 2 at the first step and the unseen path in the following steps will be overestimated significantly by IS or self-normalized IS. Thus the offline policy optimization will prefer taking action 2 at the first step as a result of overfitting. 
\end{example}

%Again we see the policy tries to avoid the logged actions on the sick patients, such that these patients contribute no weights in the weighted return policy evaluation. However, as a result of the previous action, these sick patients are not avoidable in the true environment. Thus this way of optimizing the weights is an overfitting in the offline dataset. 

Now we add a slight change in the transitions shown in Figure \ref{fig:overfitting_example}. We can see that model/value-based approach will also fail.

\begin{example}
\label{example:rl_model_fail}
In this example, we add another action in the first step. The action $3$ and action $1$ will lead to the same next state. However in the next state, no matter which action taken, the reward will depends on the action taken in the last step: If $a_1 = 1$, then we have the same reward for $a=1$ in the example in Figure \ref{fig:overfitting_example}. If $a_1 = 3$ then we have a reward $-5$. Thus model and value based method will mix the reward for $a_1 = 1$ and $a_1 = 3$ so fail in this example. Other method is not affected by the additional structure as it only add an action with minimum reward.
\end{example}

% \begin{figure}[ht]
%     \centering
%     \includegraphics[width=0.5\textwidth]{example_patient_model_fail.pdf}
%     \caption{A non-Markov variant of example in Figure \ref{fig:overfitting_example}}
%     \label{fig:example_patient_model_fail}
% \end{figure}

\clearpage
\section{Proofs of Section \ref{sec:overfitting_theory} }
\label{appendix:theory}
Proof of Theorem \ref{thm:weights_lowerbound}.
\begin{proof}
\begin{align}
    \sum_{ x_{h}^{(j)} \in \Bcal(\ith{x_h},\threshold) } \frac{\pi(a_{h}^{(j)}|x_{h}^{(j)})}{\mu(a_{h}^{(j)}|x_{h}^{(j)})} \ge& \sum_{ x_{h}^{(j)} \in \Bcal(\ith{x_h},\threshold) } \pi(a_{h}^{(j)}|x_{h}^{(j)}) \\
    \ge&  \sum_{ x_{h}^{(j)} \in \Bcal(\ith{x_h},\threshold) } \max \{0, \pi(a_{h}^{(j)}|x_{h}^{(i)}) - \threshold L \} \\
    \ge& \sum_{a \in \Aset_h(\ith{x_h};\Dcal,\threshold)} \max \{0, \pi(a|x_{h}^{(i)}) - \threshold L \} = 1 - \threshold L |\Acal|
\end{align}
\end{proof}

Proof of Corollary \ref{cor:onestep_weights_lowerbound}.
\begin{proof}
\begin{align}
     \sum_{x_{1}^{(j)} \in \Bcal(\ith{x_1},\threshold)} \frac{\max\{\ith{W},M \}}{
     \sum_{i=1}^n \max\{ \ith{W},M \} } &\ge \sum_{x_{1}^{(j)} \in \Bcal(\ith{x_1},\threshold)} \frac{\max\{\ith{W},M \}}{
     nM } \\
     &\ge \frac{\max\{\sum_{x_{1}^{(j)} \in \Bcal(\ith{x_1},\threshold)} \ith{W},M \}}{
     nM }\\ 
     &\ge \frac{1-\threshold L |\Acal|}{nM}
\end{align}
\end{proof}

Proof of Proposition \ref{prop:nstep_necessity}.
\begin{proof}
This is due to $\pi(a|\ith{x_h})$ and $\mu(a|\ith{x_h})$ are independent from history given $\ith{x_h}$. So $W_{1:h}^{(i)}$ and $\ith{W_h}$ are conditionally independent given $x_{h}^{(i)}$.
\end{proof}

Proof of Theorem \ref{thm:consistV}.
\begin{proof}
Let $\Prob_h(x;\mu) $ to be the distribution of context at $h$-th step with roll-in policy $\mu$. For any fixed $a$, we can define the distribution $\Prob_h(x|a;\mu) = \mu(a|x)\Prob_h(x;\mu)/\sum_{a} \mu(a|x)\Prob_h(x;\mu) $. For $a$ such that $\mu(a|x) > 0$, $\Prob_h(x|a;\mu)$ is also greater than zero. All $\ith{x_h}$ with $\ith{a_h} = a$ are i.i.d. samples draw from the distribution $\Prob_h(x;\mu)$. By the property of nearest neighbor \citep{cover1967nearest}, with probability 1: $$\min_{\ith{x_h} s.t. \ith{a_h} = a} \dist(x,\ith{x_h}) \to 0 < \threshold. $$
That means with probability $1$ $a \in \Aset_h(x;\Dcal,\threshold)$ for all $a$ such that $\mu(a|x) > 0$. Thus we proved the theorem statement and that the policy class will contain all $\pi$ such that $\pi(a|x) > 0$ if $\mu(a|x) > 0$. % 
%Now we construct a set of context $\{x_h^{(j)} s.t.(x_h^{(j)}, a) \in \Dcal \}$
\end{proof}

Proof of Theorem \ref{thm:consistency}.
\begin{proof}
Given the overlap assumption and Theorem \ref{thm:consistV}, for all $\pi$ we have $a \in \Aset_h(x;\Dcal,\threshold)$ for all $a$ such that $\pi(a|x) > 0$ with probability 1. Thus the solution to Equation \ref{eq:poela_objective} is the same as $\argmax_{\pi} J(\pi,\Dcal) := \hat{\pi}_{J,\Dcal}$.

By the condition that $M \to \infty$ and $\frac{M}{n} \to 0$ as $n \to \infty$, we have that the truncated IS estimator is mean square consistent \citep{ionides2008truncated}: %\sum_{i=1}^n \min\left\{   \prod_{h=1}^{H} \ith{W_h}, M \right\}
\begin{align}
    \frac{1}{ n } \sum_{i=1}^n \left( \sum_{h=1}^H \ith{r_h} \right) \min\left\{   \prod_{h=1}^{H} \ith{W_h}, M \right\}  \xrightarrow[]{q.m.} v^\pi, 
\end{align}
as $n \to \infty$. Similarly, we have that the mean of weights converge to $1$ in quadratic mean:
\begin{align}
   \frac{1}{n}\sum_{i=1}^n \min\left\{   \prod_{h=1}^{H} \ith{W_h}, M \right\} \xrightarrow[]{q.m.} 1.
\end{align}
By continuous mapping theorem, we have that the self-normalized truncated IS converge to $v^\pi$ in probability $ \wtis \xrightarrow[]{p} n  $.
The empirical variance penalty, also converge to $0$ almostly surely, since $M/n$ converge to $0$:
\begin{align}
    \frac{\sum_{i=1}^n \left(\ith{r} - \wtis \right)^2 (\min\{\ith{W},M \})^2 }{\left( \sum_{i=1}^n  \min\{\ith{W},M \}\right)^2} \le \frac{M^2}{\left( \sum_{i=1}^n  \min\{\ith{W},M \}\right)^2} \xrightarrow[]{q.m.} 0.
\end{align}
Thus the objective function $J(\pi;\Dcal)$ converge to $v^\pi$ in probability:
\begin{align}
    \Pr \left( |J(\pi;\Dcal) - v^\pi| > \epsilon \right) = \delta_n \to 0.
\end{align}
Since we assume $|\Pi|<\infty$, we have
\begin{align}
    \Pr \left( \forall \pi \in \Pi \, |J(\pi;\Dcal) - v^\pi| > \epsilon \right) = |\Pi|\delta_n. 
\end{align}
So with probability $|\Pi|\delta_n $, for any $\epsilon$:
\begin{align}
    v^{\hat{\pi}_{J,\Dcal}} \ge J(\hat{\pi}_{J,\Dcal}, \Dcal) - \epsilon \\
    \ge J(\pi^{\star}, \Dcal) - \epsilon \\
    \ge v^{\pi^{\star}} - 2\epsilon, 
\end{align}
where $\pi^\star$ is $\argmax_{\pi \in \Pi} v^\pi $. As $|\Pi|\delta_n \to 0$, we proved the true value of empirical maximizer $v^{\hat{\pi}_{J,\Dcal}}$ converge to the maximum of value $\max_{\pi \in \Pi} v^\pi$ in probability.
\end{proof}

\section{Experiment Details}
\label{appendix:experiment-details}
\textbf{For all experiments in the main text}, we report the test performance of the policy saved at the end of training either through online Monte-Carlo estimation if a simulator is available, or using SNTIS estimates on a held out test set.

\begin{figure}[ht]
    \centering
    \includegraphics[width=0.9\textwidth]{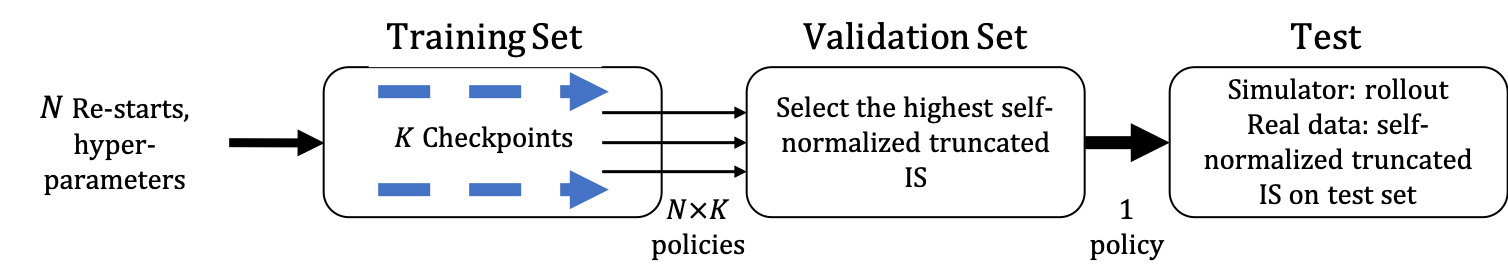}
    \caption{The process of hyperparameters search and test in the experiment.}
    \label{fig:exp_flow}
\end{figure}

\textbf{For all experiments reported in Appendices~\ref{ap:tumor-checkpoint} and~\ref{ap:sepsis-checkpoint}}, we follow the 3-phases pipelines we describe hereafter to decide the test score we report in the corresponding Tables. To put ourselves in the more realistic situation of real-world applications where practitioners would select a policy from regular checkpoints along its training on the basis of its SNTIS score on the validation set, an algorithm is trained on the training set multiple times, using different hyperparameters and several restarts. Intermittent policies generated during the training process identified with the highest self-normalized truncated IS (SNTIS) estimates on a held-out validation set are saved at checkpoints. The pipeline is illustrated in Figure \ref{fig:exp_flow}.

The open-source code for \ispg can be found here: \href{https://github.com/StanfordAI4HI/poela}{https://github.com/StanfordAI4HI/poela}.

\subsection{Experiment Details in TGI Simulator}
The TGI simulator describes low-grade gliomas (LGG) growth kinetics in response to chemotherapy in a horizon of 30 months using an ordinary differential equation model. The parameter in ODEs are estimated using data from adult diffuse LGG during and after chemotherapy was used, in a horizon of 30 months. The goal in this environment is to achieve a reduction in mean tumor diameters (MTD) while reducing the drug dosage \citep{yauney2018reinforcement}. We includes the MTD, the drug concentration, and the number of month (time-step) in the context space. Notice that this context space is non-Markov as it does not include all parameters in the ODEs. Actions are binary representing taking the full dose or no dose which is same as prior work \citep{yauney2018reinforcement}. The reward at each step consist of an immediate penalty proportional to the drug concentration, and a delayed reward at the end measures the decrease of MTD compared with the beginning. Each episodes, the parameters including the initial MTD are sampled from a log-Normal distribution as \citep{ribba2012tumor} representing the difference in individuals. The behavior policy is a fixed dosing schedule of 9 months (the median duration from \cite{peyre2010prolonged}) plus $30\%$ of a uniformly random choice of actions. We run all algorithms on a training set with 1000 episodes with different hyperparameters (listed below), and 5 restarts, saving checkpoints along the training.% Then we select the best policy for each algorithm by $\wtis$ on the validation set with 1000 episodes as well.
The validation set is comprised of 1000 episodes as well. 

\paragraph{Hyperparameters.} In the first part of Table \ref{tab:hp_tumor} we show the searched hyperparameters of each algorithm, except that the parameter $b$ in PQL is set adaptively as the $2$-percentile of the score on the training set as in the original paper \cite{liu2020provably}. As we know the behavior policy, we use the true behavior policy in BCQ and PQL algorithm. So BCQ threshold takes only two values as the behavior policy is $\epsilon$-deterministic so there are only two distinct values. In the second part of Table \ref{tab:hp_tumor} we specify some fixed hyperparameters/settings for all algorithm. All policy/Q functions are approximated by fully-connected neural networks with two hidden layers with 32 units.

\begin{table}[ht]
    \centering
    \begin{tabular}{c|cc}
    \toprule
       Hyperparameters  & used in algorithms & values  \\
       \midrule
       $\delta$ & \ispg & $0.05, 0.1, 0.5$ \\
        $\hat{\mu}$ threshold & \ispgknn & 0.01, 0.05, 0.1, 0.2 \\
       CRM Var coefficient & \ispg, \ispgbaseline & $0, 0.1, 1$ \\
       BCQ threshold & BCQ, PQL & $0.0, 0.2$ \\ 
       \midrule
       $M$ in $\wtis$ & \text{All} & $1000$ \\
       Max training steps & \ispg, \ispgbaseline, \ispgknn & 500 \\
       & BCQ, PQL & 1000 \\
       Number of checkpoints & All & 50 \\
       Batch size & BCQ, PQL & 100 \\
       \bottomrule
    \end{tabular}
    \vspace{0.5cm}
    \caption{Hyperparameters in the TGI simulator experiment}
    \label{tab:hp_tumor}
\end{table}
The difference in the max update steps and checkpoints frequency is caused by the fact that BCQ and PQL is updated by stochastic gradient descent and all policy optimization based on SNTIS is using gradient descent.

% A single training and validation run of an algorithm with takes no more than half an hours on Intel Xeon CPU with 2.40 GHz.

\subsection{Experiment Details in the MIMIC III Dataset}
\label{ap:sepsis-details}
The MIMIC III sepsis dataset is available upon application and training: https://mimic.mit.edu/iii/gettingstarted/. The code to extract the cohort is available on: https://gitlab.doc.ic.ac.uk/AIClinician/AIClinician. This cohort consists of data for 14971 patients. The contexts for each patient consist of 44 features, 
%including demographics, Elixhauser premorbid status, vital signs, laboratory values,
summarized in 4-hour intervals, for at most 20 steps. The actions we consider are the prescription of IV fluids and vasopressors. Each of the two treatments is binned into 5 discrete actions according to the dosage amounts, resulting in 25 possible actions. The rewards are defined from the 90-day mortality in the logs, $100$ if the patient survives  and $0$ otherwise.

We now provide details of the experiment on MIMIC III sepsis dataset here. We run all algorithms on a training set with 8982 trajectories with different hyperparameters (listed below), and 3 restarts, saving checkpoints along the training.% Then we select the best policy for each algorithm by $\wtis$ on the validation set with 2994 trajectories.
The validation set is comprised of 2994 trajectories. Finally we get the $\wtis$ evaluation on the test set with 2995 trajectories. In the first part of Table~\ref{tab:hp_sepsis} we list the hyperparameters that we searched on the validation set for each algorithm, except that the parameter $b$ in PQL is set adaptively as the $2$-percentile of the score on the training set as in the original paper~\citep{liu2020provably}. In the second part of Table~\ref{tab:hp_tumor} we specify some fixed hyperparameters/settings for all algorithm. All policy/Q-functions are approximated by fully-connected neural networks with two hidden layers with 256 units.

\begin{table}[ht]
    \centering
    \begin{tabular}{c|cc}
    \toprule
       Hyperparameters  & used in algorithms & values  \\
       \midrule
       $\delta$ & \ispg & $0.4, 0.6, 0.8, 1.0$ \\
       $\hat{\mu}$ threshold & \ispgknn & 0.01, 0.02, 0.05, 0.1 \\
       CRM Var coefficient & \ispg, \ispgbaseline & $0, 0.1, 1, 10$ \\
       BCQ threshold & BCQ, PQL & $0.0, 0.01, 0.05, 0.1, 0.3, 0.5$ \\ 
       \midrule
       $M$ in $\wtis$ & \text{All} & $1000$ \\
       Max training steps & \ispg, \ispgbaseline, \ispgknn & 1000 \\
       & BCQ, PQL & 10000 \\
       Number of checkpoints & All & 100  \\
       Batch size & BCQ, PQL & 100 \\
       \bottomrule
    \end{tabular}
    \vspace{0.5cm}
    \caption{Hyperparameters in the MIMIC III sepsis experiment}
    \label{tab:hp_sepsis}
\end{table}

As we explained, the difference in the max update steps and checkpoints frequency is caused by the fact that BCQ and PQL is updated by stochastic gradient descent and all policy optimization based on SNTIS is using gradient descent.

% A single training and validation run of an algorithm with takes no more than 2 hours on Intel Xeon CPU with 2.40 GHz.

\subsection{Experiment Details for the Behavior policy Estimation}
\label{ap:bc-details}
In the implementation of BC, we use Multi-Layer Perceptrons (MLPs) neural networks with layer dimensions [32, 32, 32] for the LGG Tumor Growth Inhibition simulator and [256, 256, 256] for the MIMIC III dataset. All use ReLU activations. For BCRNN, we use 3-layer GRUs with a RNN hidden dimension of size 100. All networks are trained using Adam optimizer~\citep{kingma2014adam} with learning rate $3e-4$. For all experiments, BC and BCRNN are trained for 500 steps and directly serve as estimated behavior policies.

\subsection{Importance weights in low-reward trajectories}
\label{sec:is_low}
To examine if the proposed overfitting phenomenon exists in real experimental datasets, we compute the importance weights of the learned policy on the low-reward  trajectories in the training data for our MIMIC III dataset and our tumor simulator. Our hypothesis is that overfitting of the importance weights in policy gradient methods may result in the algorithm avoiding initial states with low rewards, which motivated our proposed algorithm. 
%We show this on the training set to reflect our hypothesis of overfitting importance weights in policy gradient methods. 

%For the completeness of this results and ablation study reason, we also shows this for the batch Q learning baselines (BCQ and PQL). Though the Q learning methods might not overfit the importance weights. 

In MIMIC III dataset the reward for a trajectory is either $0$ or $100$. We define the low-reward trajectories as those with $0$ reward. Low-reward trajectories are over $60\%$ of all trajectories in the dataset. In the Tumor simulation experiment we define a  low-reward trajectory  when reward is less than $-2$. Over $95\%$ of trajectories in the Tumor simulation dataset are low-reward trajectories. 

%The hyperparameters for each algorithm is the same as the selected one in the main experimental section. %**EB: we are only evaluating the learned policy on these states so I think the hyperparameters should be the same by default

The table below shows, for each algorithm and setting, the sum of the SNTIS weights of the learned policy on the training set, for low-reward trajectory states. Our primary interest is to illustrate that alternate policy gradient methods that are also suitable for non-Markov domains, can exhibit the importance sampling overfitting of avoiding low reward trajectories. We indeed see in Table~\ref{tab:training_weights} that POELA has a much larger weight on low-reward trajectories than  alternate offline policy search methods: 

\begin{table}[tbh]
    \centering
    \begin{tabular}{c|ccc}
        \toprule 
         Method & \ispg& \ispgknn & \ispgbaseline  \\
         \midrule 
        MIMIC III & 0.028 & 0.001 & 0.003 \\%& 0.148 & 0.149 \\
        Tumor non-MDP & 0.054 & - (fixed policy) & 0.005 \\ %& 0.0003 & 0.005 \\
        %Tumor MDP & 0.097 & - (fixed policy) & 0.0004 & 0.083 & 0.124 \\
         \bottomrule
    \end{tabular}
    \vspace{0.1cm}
    \caption{ Importance weights overfitting: sum of SNTIS weights of learned policy on the training set. }
    \label{tab:training_weights}
\end{table}

The Q-learning baselines we consider (BCQ and PQL) do not directly use the importance weights, but they do try to avoid actions and/or states and actions with little support. Our POELA method can be viewed as being similarly inspired, but for non-Markovian settings where policy gradient is beneficial. We also compute the SNTIS weights of the BCQ/PQL policy on the training set in the Markov domain that satisfies the Markov assumptions of BCQ/PQL. In Table~\ref{tab:training_weights_mdp} we can see that  POELA, BCQ and PQL all still give significantly more weight to low reward trajectories than the alternate policy gradient methods:

\begin{table}[tbh]
    \centering
    \begin{tabular}{c|ccccc}
        \toprule 
         Method & \ispg& \ispgknn & \ispgbaseline & BCQ  & PQL \\
         \midrule 
        Tumor MDP & 0.097 & - (fixed policy) & 0.0004 & 0.083 & 0.124 \\
         \bottomrule
    \end{tabular}
    \vspace{0.1cm}
    \caption{ Importance weights overfitting: sum of SNTIS weights of learned policy on the training set. }
    \label{tab:training_weights_mdp}
\end{table}

%\begin{table}[tbh]
%\color{red}
 %   \centering
 %   \begin{tabular}{c|ccccc}
  %      \toprule 
%         Method & \ispg& \ispgknn & \ispgbaseline & BCQ  & PQL \\
 %        \midrule 
  %      MIMIC III & 0.028 & 0.001 & 0.003 & 0.148 & 0.149 \\
   %     Tumor non-MDP & 0.054 & - (fixed policy) & 0.005 & 0.0003 & 0.005 \\
%        Tumor MDP & 0.097 & - (fixed policy) & 0.0004 & 0.083 & 0.124 \\
 %        \bottomrule
 %   \end{tabular}
 %   \vspace{0.1cm}
 %   \caption{\color{red} Importance weights overfitting: sum of SNTIS weights of learned policy on the training set. }
 %   \label{tab:training_weights}
%\end{table} 

These results help illustrate that the over avoidance of low-reward trajectories can be observed by past policy gradient methods in our datasets. Of course, one challenge is that in real settings, an excellent policy may have low importance weights in avoidable low-reward states and trajectories, but should have higher importance weights in non-avoidable low reward starting states and trajectories. To get a fuller picture of performance, it is helpful to look both at the weights on trajectories with low rewards and the test evaluation results.  
%We also want to emphasize that the splitting of low-reward/high-reward uses the reward information. Unlike the illustrative example, in this case, we cannot know that if it really learns a better decision or just avoids a particular state which should not be affected by no matter what decisions. In fact, a really good policy will also have very low importance weights in the low-reward samples. These results need to be combined with the test evaluation results to understand the overfitting phenomenon. 
Compared with strong policy gradient baselines, our proposed regularization method have larger  importance weights on low-reward trajectories, and the gap between training/validation evaluation and online test performance is also smaller, suggesting that we are less likely to learn policies that erroneously believe they can avoid unavoidable low reward settings. 

\subsection{The effect of eligible action constraints $\delta$}
\label{sec:effect_delta}
In this section we explore how the choice of $\delta$, which constrains the policy class through impacting the eligible actions, impacts empirical performance. Larger $\delta$ corresponds to a less constrained policy class.  %To show how much the proposed eligible action constraints on actions affect the performance in practice. We shows an ablation study of the training and test SNTIS score given hyper-parameter $\delta$'s value. The higher $\delta$ is, the less constraints we make on actions. 
Other hyperparameters are selected by the same procedure as described in previous sections.

Table~\ref{tab:sepsis_delta_study} shows the results. As $\delta$ increases, the policy search operates with less constraints. The results show that in this case, our 
policy gradient method produces a policy with a higher value in the training set, but that policy may not perform as well in the test evaluation, and may have a smaller effective sample size than when a smaller $\delta$ is used. 
The best hyperparameter value $\delta$ lies in the middle of the explored range. $\delta$ can be selected based on  performance and effective sample size.

\begin{table}[tbh]
    \centering
    \begin{tabular}{l|ccccc}
        \toprule 
         $\delta$ & 0.4& 0.6 & 0.8 & 1.0 \\
         \midrule 
        training $\wtis$ & 91.62 & 98.41 & 98.9 & 99.12 \\
        training ESS & 3601.12 & 2242.07 & 1993.08 &1769.46\\
        test $\wtis$ & 86.62 & 90.07 & 91.46 & 90.23  \\
        test ESS  & 1278.08 & 819.64 & 624.92 & 542.53 \\
         \bottomrule
    \end{tabular}
    \vspace{0.1cm}
    \caption{ The effect of eligible action constraints $\delta$ on the results in MIMIC III sepsis dataset. }
    \label{tab:sepsis_delta_study}
\end{table}

\clearpage
\section{Additional experiments: LGG Tumor Growth Inhibition simulator}
\label{ap:more-exp-tumor}
In this section, we provide additional experiments to the existing LGG Tumor Growth Inhibition simulator experiments.

\subsection{Experiment with estimating the behavior policy with function approximation}
\label{ap:tumor-bc-exp}
\begin{table*}[ht]
    \centering
    \begin{small}
    \tabcolsep=0.1cm % reduce table size by reducing column separation
    \begin{tabular}{c|c|ccccc|c}
        \toprule 
         & Algorithms & \ispg & \ispgmuhat & \ispgbaseline & BCQ & PQL & $9$-mon \\
         \midrule 
         Non-MDP & Test $v^\pi$ & $ 92.34 \pm 1.57 $ & $ 59.62 \pm 12.71 $ & $ 46.66 \pm 14.05$ &  $ 19.36 \pm 5.66 $ &  $30.44 \pm 10.38 $& $68.12 $ \\
         & $\wtis - v^\pi$ & $ 0.94 \pm 1.66 $ & $31.38 \pm 10.97$ & $42.98 \pm 12.87$ &  $72.35 \pm 5.66$&  $62.24 \pm 10.94$ & $-$ \\
        \midrule
        MDP & Test $v^\pi$ & $91.04 \pm 0.55 $ & $78.21 \pm 4.94 $ & $78.70 \pm 0.60 $ &  $99.26 \pm 0.59 $ &  $99.66 \pm 0.29 $ & $68.12$ \\
        & $\wtis - v^\pi$ & $3.40 \pm 2.48 $ & $15.58 \pm 3.92 $ & $15.10 \pm 3.97 $ &  $ -3.88 \pm 1.60 $&  $ -4.09 \pm 1.75 $ & $-$ \\
         \bottomrule
    \end{tabular}
    \end{small}
    \caption{LGG Tumor Growth Inhibition simulator. Test $v^\pi$ and amount of overfitting of the learned policy. Test $v^\pi$ is computed from 1000 rollouts in the simulator. $\wtis$ on the validation set $-$ test $v^\pi$ represents the amount of overfitting. %Drug dosing and treatment effect measured by MTD change is computed from the test rollouts. 
    All numbers are averaged across 5 runs with the standard error reported. Behavior policy $\hat{\mu} = \text{BC}$.}
    \label{tab:tumor_result_bc}
    \vspace{-0.3cm}
\end{table*}

\begin{table*}[ht]
    \centering
    \begin{small}
    \tabcolsep=0.1cm % reduce table size by reducing column separation
    \begin{tabular}{c|c|ccccc|c}
        \toprule 
         & Algorithms & \ispg & \ispgmuhat & \ispgbaseline & BCQ & PQL & $9$-mon \\
         \midrule 
         Non-MDP & Test $v^\pi$ & $ 95.81 \pm 1.68 $ & $76.64 \pm 14.65 $ & $76.43 \pm 14.59 $ &  $ 19.79 \pm 5.76  $ &  $ 31.57 \pm 10.63   $ & $68.12 $ \\
         & $\wtis - v^\pi$ & $-1.52 \pm 1.79 $ & $16.35\pm 14.40 $ & $16.56\pm 14.36 $ &  $ 73.71 \pm 6.34   $ &  $ 62.92  \pm 11.19   $ & $-$ \\
        %  Drug dosing & $5.31 \pm 0.31 $ & $3.81 \pm 1.00$ & $\textbf{0.80} \pm 0.72$ & $2.56 \pm 0.56$ & $9$ \\
        \midrule
        MDP & Test $v^\pi$ & $89.25 \pm 1.51 $ & $75.43 \pm 8.25 $ & $73.61 \pm 0.30 $ &  $ 99.57\pm0.29  $ &  $99.96 \pm 0.12  $ & $68.12$ \\
        & $\wtis - v^\pi$ & $5.17 \pm 2.20 $ & $17.66 \pm 7.90  $ & $19.44 \pm 8.52 $ &  $-4.18 \pm 1.76  $ & $ -4.38 \pm 1.78  $ & $-$ \\
         \bottomrule
    \end{tabular}
    \end{small}
    \caption{LGG Tumor Growth Inhibition simulator. Test $v^\pi$ and amount of overfitting of the learned policy. Test $v^\pi$ is computed from 1000 rollouts in the simulator. $\wtis$ on the validation set $-$ test $v^\pi$ represents the amount of overfitting. %Drug dosing and treatment effect measured by MTD change is computed from the test rollouts. 
    All numbers are averaged across 5 runs with the standard error reported. Behavior policy $\hat{\mu} = \text{BCRNN}$.}
    \label{tab:tumor_result_bcrnn}
    \vspace{-0.3cm}
\end{table*}

\subsection{Alternative selection procedure: checkpoint best intermittent policies}
\label{ap:tumor-checkpoint}
In this section, we use the procedure of best policy checkpoint during the training described in Section~\ref{appendix:experiment-details}. We report the test performance of the selected policy through online Monte-Carlo estimation.

\begin{table*}[h!]
    \centering
    \begin{small}
    \tabcolsep=0.1cm % reduce table size by reducing column separation
    \begin{tabular}{c|c|ccccc|c}
        \toprule 
         & Algorithms & \ispg & \ispgknn & \ispgbaseline & BCQ & PQL & $9$-mon \\
         \midrule 
         Non-MDP & Test $v^\pi$ & $ 92.20 \pm 1.63 $ & $76.99 \pm 13.80$ & $75.06 \pm 13.22$ & $57.77 \pm 16.71$  & $74.76 \pm 9.75$  & $68.12 $ \\
         & $\wtis - v^\pi$ & $ -1.26 \pm 1.92$ & $16.07 \pm 13.55$ & $15.57 \pm 13.07$ & $37.55 \pm 16.91$  & $17.74 \pm 9.49$ & $-$ \\
        %  Drug dosing & $5.31 \pm 0.31 $ & $3.81 \pm 1.00$ & $\textbf{0.80} \pm 0.72$ & $2.56 \pm 0.56$ & $9$ \\
        \midrule
        MDP & Test $v^\pi$ & $89.52 \pm 1.55$ & $69.18 \pm 10.17$ & $78.79 \pm 6.42$ & $94.7\pm3.49$  & $96.88 \pm 3.76$  & $68.12$ \\
        & $\wtis - v^\pi$ & $ 5.16 \pm 1.78$ & $24.92 \pm 9.71$ & $14.93 \pm 5.71$ & $2.75 \pm 3.41$  & $-0.26 \pm 4.18$  & $-$ \\
         \bottomrule
    \end{tabular}
    \end{small}
    \caption{LGG Tumor Growth Inhibition simulator. Test $v^\pi$ and amount of overfitting of the learned policy. Test $v^\pi$ is computed from 1000 rollouts in the simulator. $\wtis$ on the validation set $-$ test $v^\pi$ represents the amount of overfitting. %Drug dosing and treatment effect measured by MTD change is computed from the test rollouts. 
    All numbers are averaged across 5 runs with the standard error reported. \textbf{Procedure: best intermittent policy checkpoints.}}
    \label{tab:tumor_result_checkpoint}
    \vspace{-0.3cm}
\end{table*}

\begin{table*}[h!]
    \centering
    \begin{small}
    \tabcolsep=0.1cm % reduce table size by reducing column separation
    \begin{tabular}{c|c|ccccc|c}
        \toprule 
         & Algorithms & \ispg & \ispgmuhat & \ispgbaseline & BCQ & PQL & $9$-mon \\
         \midrule 
         Non-MDP & Test $v^\pi$ & $ 94.16 \pm 1.82 $ & $ 74.76 \pm 7.66 $ & $ 76.38 \pm 7.26$ & $92.92 \pm 1.68 $  & $74.65 \pm 14.5 $ & $68.12 $ \\
         & $\wtis - v^\pi$ & $ 0.95 \pm 1.92 $ & $18.02 \pm 7.07$ & $15.02 \pm 6.68$ & $0.58 \pm 0.27$ & $20.49 \pm 14.46$  & $-$ \\
        \midrule
        MDP & Test $v^\pi$ & $91.81 \pm 1.05 $ & $84.86 \pm 3.48 $ & $84.08 \pm 3.46 $ & $86.22 \pm 9.61 $  & $95.02 \pm 4.95 $ & $68.12$ \\
        & $\wtis - v^\pi$ & $2.7 \pm 2.82 $ & $9.23 \pm 3.85 $ & $10.01 \pm 3.88 $ & $11.03 \pm 10.4 $ & $2.45 \pm 5.27 $  & $-$ \\
         \bottomrule
    \end{tabular}
    \end{small}
    \caption{LGG Tumor Growth Inhibition simulator. Test $v^\pi$ and amount of overfitting of the learned policy. Test $v^\pi$ is computed from 1000 rollouts in the simulator. $\wtis$ on the validation set $-$ test $v^\pi$ represents the amount of overfitting. %Drug dosing and treatment effect measured by MTD change is computed from the test rollouts. 
    All numbers are averaged across 5 runs with the standard error reported. Behavior policy $\hat{\mu} = \text{BC}$. \textbf{Procedure: best intermittent policy checkpoints.}}
    \label{tab:tumor_result_bc_checkpoint}
    \vspace{-0.3cm}
\end{table*}

\begin{table*}[h!]
    \centering
    \begin{small}
    \tabcolsep=0.1cm % reduce table size by reducing column separation
    \begin{tabular}{c|c|ccccc|c}
        \toprule 
         & Algorithms & \ispg & \ispgmuhat & \ispgbaseline & BCQ & PQL & $9$-mon \\
         \midrule 
         Non-MDP & Test $v^\pi$ & $ 96.34 \pm 1.58 $ & $77.51 \pm 13.87 $ & $75.73 \pm 14.3 $ & $92.73 \pm 1.67 $  & $74.94 \pm 14.47 $  & $68.12 $ \\
         & $\wtis - v^\pi$ & $-2.05 \pm 1.9 $ & $15.48\pm 13.62 $ & $17.27\pm 14.02 $ & $0.77\pm 0.52 $  & $20.2\pm 14.43 $  & $-$ \\
        %  Drug dosing & $5.31 \pm 0.31 $ & $3.81 \pm 1.00$ & $\textbf{0.80} \pm 0.72$ & $2.56 \pm 0.56$ & $9$ \\
        \midrule
        MDP & Test $v^\pi$ & $90.06 \pm 1.65 $ & $79.62 \pm 7.82 $ & $79.54 \pm 7.65 $ & $86.38 \pm 9.47 $  & $95.16 \pm 4.9 $  & $68.12$ \\
        & $\wtis - v^\pi$ & $4.46 \pm 2.31 $ & $13.81 \pm 6.96  $ & $13.89 \pm 6.81 $ & $10.87 \pm 10.24 $  & $2.33 \pm 5.19 $  & $-$ \\
         \bottomrule
    \end{tabular}
    \end{small}
    \caption{LGG Tumor Growth Inhibition simulator. Test $v^\pi$ and amount of overfitting of the learned policy. Test $v^\pi$ is computed from 1000 rollouts in the simulator. $\wtis$ on the validation set $-$ test $v^\pi$ represents the amount of overfitting. %Drug dosing and treatment effect measured by MTD change is computed from the test rollouts. 
    All numbers are averaged across 5 runs with the standard error reported. Behavior policy $\hat{\mu} = \text{BCRNN}$. \textbf{Procedure: best intermittent policy checkpoints.}}
    \label{tab:tumor_result_bcrnn_checkpoint}
    \vspace{-0.3cm}
\end{table*}

\section{Additional experiments: MIMIC III sepsis}
\label{ap:more-exp-sepsis}

In this section we provide additional experiments to the existing MIMIC III sepsis experiments.

\subsection{Alternative selection procedure: checkpoint best intermittent policies}
\label{ap:sepsis-checkpoint}
In this section, we use the procedure of using checkpoints to select best policies during the training described in Section~\ref{appendix:experiment-details}. We report the test performance of the selected policy using SNTIS estimates on a held out test set.

\begin{table*}[h!]
    \centering
    \begin{small}
    \begin{tabular}{c|ccccc|c}
        \toprule 
         Method & \ispg& \ispgmuhat & \ispgbaseline & BCQ  & PQL & Clinician \\
         \midrule 
        Test SNTIS & 91.46 (90.82) & 87.95 & 87.71 & 82.67 & 84.40 & 81.10 \\
        $95\%$ BCa UB & 93.24 (92.61) &	90.58 &		90.04 & 86.83 &	88.29 & 82.19 \\
        $95\%$ BCa LB & 89.59 (88.68) & 84.77 &	84.90 & 78.25	& 80.13 & 79.80\\
        Test ESS & 624.92 (586.37) & 372.00 & 399.59 & 228.82 & 231.93 & 2995 \\
         \bottomrule
    \end{tabular}
    \end{small}
    \caption{MIMIC III sepsis dataset. Test evaluation, $(0.05, 0.95)$ BCa bootstrap interval, and effective sample size. The value of \ispg without a CRM variance penalty is shown in parentheses. \textbf{Procedure: best intermittent policy checkpoints.}}
    \label{tab:sepsis_result_checkpoint}
    \vspace{-0.3cm}
\end{table*}

\begin{table*}[h!]
    \centering
    \begin{small}
    \begin{tabular}{c|ccccc|c}
        \toprule 
         Method & \ispg& \ispgmuhat & \ispgbaseline & BCQ  & PQL & Clinician \\
         \midrule 
        Test SNTIS & $85.01$ ($89.62$) & $84.70$ &	$85.53$ & $83.17$& $84.16$ & 81.10 \\
        $95\%$ BCa UB & $88.61$ ($92.75$) & $88.56$ & $87.80$ & $92.88$	& $88.04$ & 82.19 \\
        $95\%$ BCa LB & $80.55$ ($85.57$) & $80.15$ & $83.23$ & $63.98$	& $79.98$ & 79.80\\
        Test ESS & $227.92$ ($214.12$) & $228.97$ &	$354.86$ & $208.92$	& $209.72$ & 2995 \\
         \bottomrule
    \end{tabular}
    \end{small}
    \caption{MIMIC III sepsis dataset. Test evaluation, $(0.05, 0.95)$ BCa bootstrap interval, and effective sample size. The value of \ispg without a CRM variance penalty is shown in parentheses. Behavior policy $\hat{\mu} = \text{BC}$. \textbf{Procedure: best intermittent policy checkpoints.}}
    \label{tab:sepsis_result_bc_checkpoint}
    \vspace{-0.3cm}
\end{table*}

\begin{table*}[h!]
    \centering
    \begin{small}
    \begin{tabular}{c|ccccc|c}
        \toprule 
         Method & \ispg& \ispgmuhat & \ispgbaseline & BCQ  & PQL & Clinician \\
         \midrule 
        Test SNTIS & $88.34$ ($90.89$) & $87.98$ & $85.12$	& $83.20$	& $85.06$ & 81.10 \\
        $95\%$ BCa UB & $91.65$ ($93.78$) & $91.06$ & $92.75$ & $91.56$	& $89.12$ & 82.19 \\
        $95\%$ BCa LB & $83.94$ ($87.05$) & $84.41$ & $72.96$ & $66.27$	& $79.76$ & 79.80\\
        Test ESS & $201.49$ ($220.86$) & $285.82$ & $211.20$ & $206.11$ & $212.36$ & 2995 \\
         \bottomrule
    \end{tabular}
    \end{small}
    \caption{MIMIC III sepsis dataset. Test evaluation, $(0.05, 0.95)$ BCa bootstrap interval, and effective sample size. The value of \ispg without a CRM variance penalty is shown in parentheses. Behavior policy $\hat{\mu} = \text{BCRNN}$. \textbf{Procedure: best intermittent policy checkpoints.}}
    \label{tab:sepsis_result_bcrnn_checkpoint}
    \vspace{-0.3cm}
\end{table*}

\subsection{Ablation study: ESS constraints for hyperparameter selection on validation set}
\label{ap:sepsis-ablation}
In the main text, we set an effective sample size threshold of 200 for a policy/hyperparameter to be selected on validation set. This is to make sure we have large enough effective sample size on the test set to provide reliable off-policy test estimates. In Table~\ref{tab:sepsis_noess_result}, we show the results if we do not threshold the effective sample size on validation set. Generally, all algorithms will prefer a high off-policy estimates without enough effective sample size. On the test set, all algorithms yields a small effective sample size, thus unreliable off-policy estimates and large bootstrap confidence interval. The proposed methods is better than baselines but also has much smaller $95\%$ bootstrap lower bound than with the effective sample size constraint.

\begin{table}[ht]
    \centering
    \begin{tabular}{c|ccccc}
        \toprule 
         Method & \ispg& \ispgknn & \ispgbaseline & BCQ  & PQL \\
         \midrule 
        Test SNTIS & 87.63(86.29) & 82.36 & 82.36 & 83.28 & 96.32 \\
        $95\%$ BCa LB & 85.06(83.51) &	64.92 &		63.48 &				56.65 &	57.25 \\
        $95\%$ BCa UB & 90.00(88.59) &	94.22 &		93.62 &				100 &	100\\
        Test ESS & 528.18(491.71) &	21.23 &		21.23 &				9.04 &	1.27 \\
         \bottomrule
    \end{tabular}
    \vspace{0.1cm}
    \caption{Test evaluation without effective sample size constraint on the validation set, $(0.05, 0.95)$ BCa bootstrap interval, and effective sample size in the sepsis cohort of MIMIC III dataset. Value inside parenthesis of \ispg is without CRM variance penalty. }
    \label{tab:sepsis_noess_result}
    \vspace{-0.5cm}
\end{table}

% \newpage
\subsection{The trade-off between ESS and performance estimates}
\label{ap:sepsis-tradeoff}
\begin{figure*}[ht]%{R}{0.8\textwidth}
    \centering
    \includegraphics[width=0.6\textwidth]{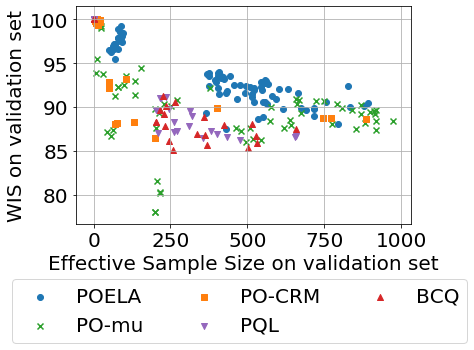}
    \caption{Trade-off between ESS and value estimates.}
    \label{fig:sepsis_tradeoff}
\end{figure*}

A tension in conservative offline optimization is that the most reliable and conservative policy estimates come from effectively imitating the behavior policy (which will maximize ESS). Policies that differ substantially from the behavior policy may yield higher performance, but have less overlap with the existing logged data (and lower ESS). This is illustrated in Figure~\ref{fig:sepsis_tradeoff}, where the value estimates are plotted for each hyperparameter and re-start of the different algorithms. We observe that \ispg achieves a better Pareto frontier between performance estimates and ESS than other algorithms. Note that for this experiment we placed ourselves in the policy selection procedure in which the best policy is selected during training based on SNTIS estimates on the validation set (cf. Table~\ref{tab:sepsis_result_checkpoint}).

\subsection{Eligible actions visualization for high/mid/low-SOFA patients}
\label{ap:sepsis-action-viz}
\begin{figure}[h!]%{R}{0.8\textwidth}
    \begin{minipage}{\textwidth}
    \centering
    \includegraphics[width=0.8\textwidth]{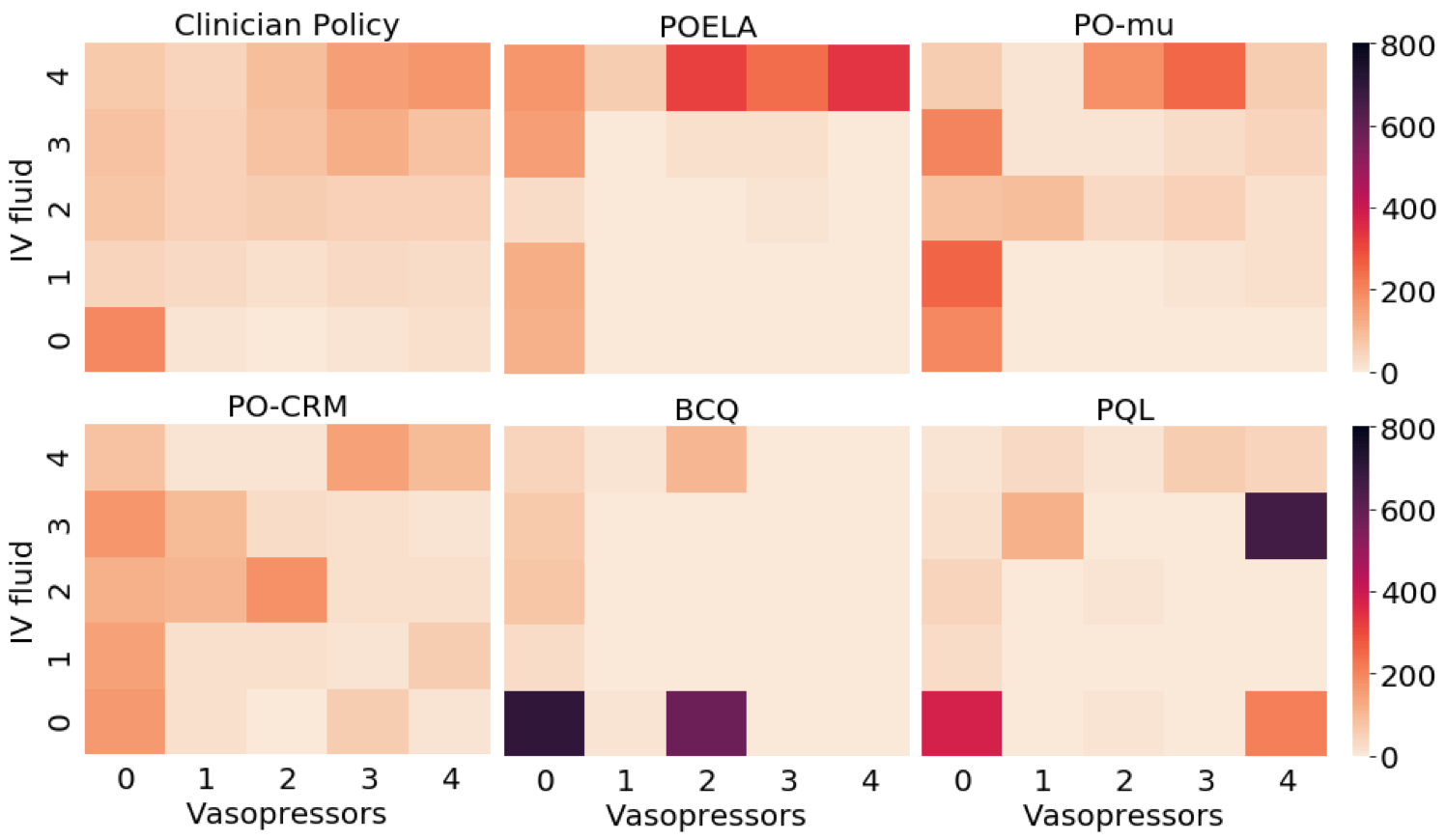}
    \subcaption{Action counts in high-SOFA contexts}
    \label{fig:sepsis_highsofa_visualization}
    \end{minipage} \\
    \begin{minipage}{\textwidth}
    \centering
    \includegraphics[width=0.8\textwidth]{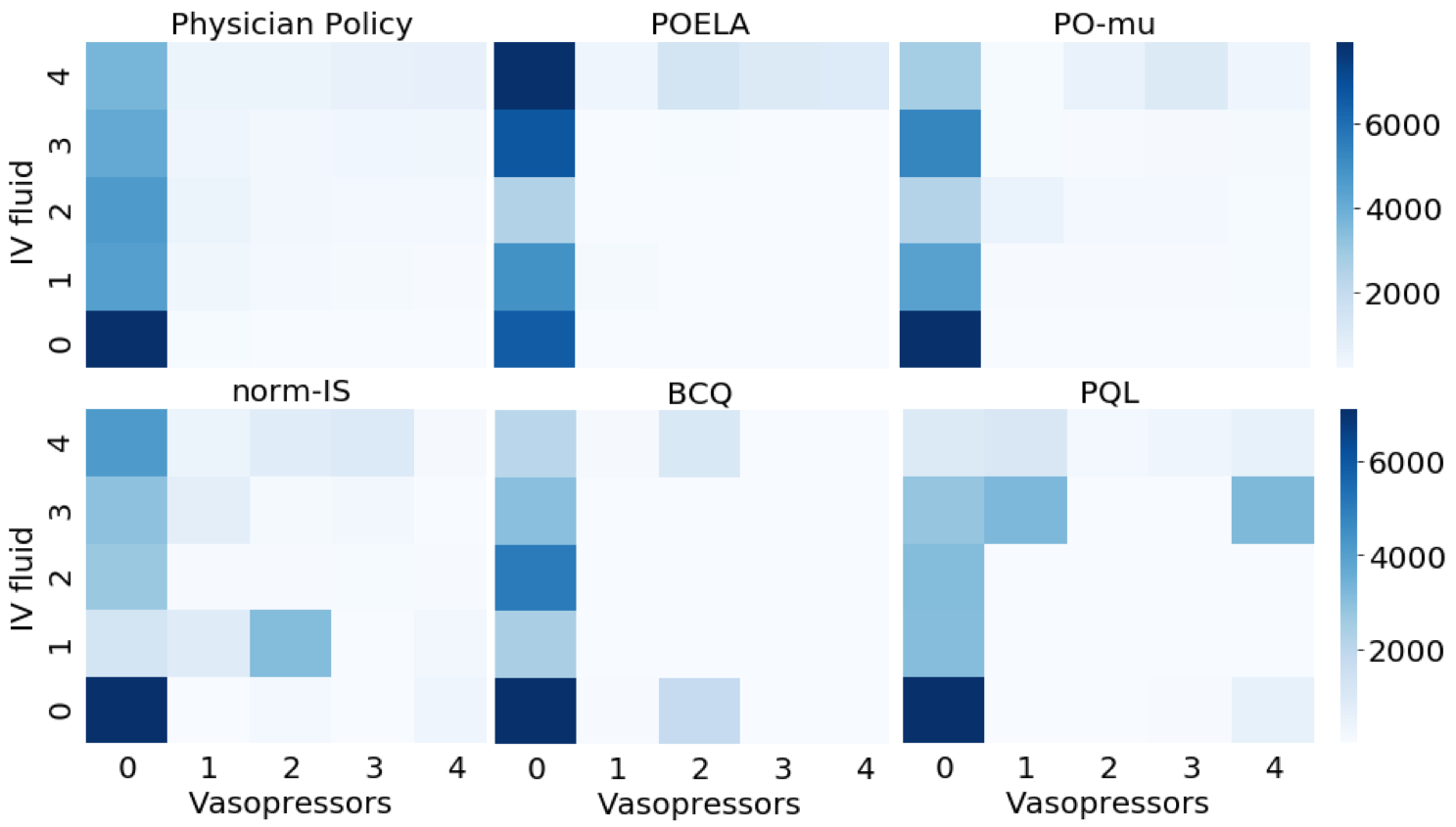}
    \subcaption{Action counts in mid-SOFA contexts}
    \label{fig:sepsis_midsofa_visualization}
    \end{minipage} \\
    \begin{minipage}{\textwidth}
    \centering
    \includegraphics[width=0.8\textwidth]{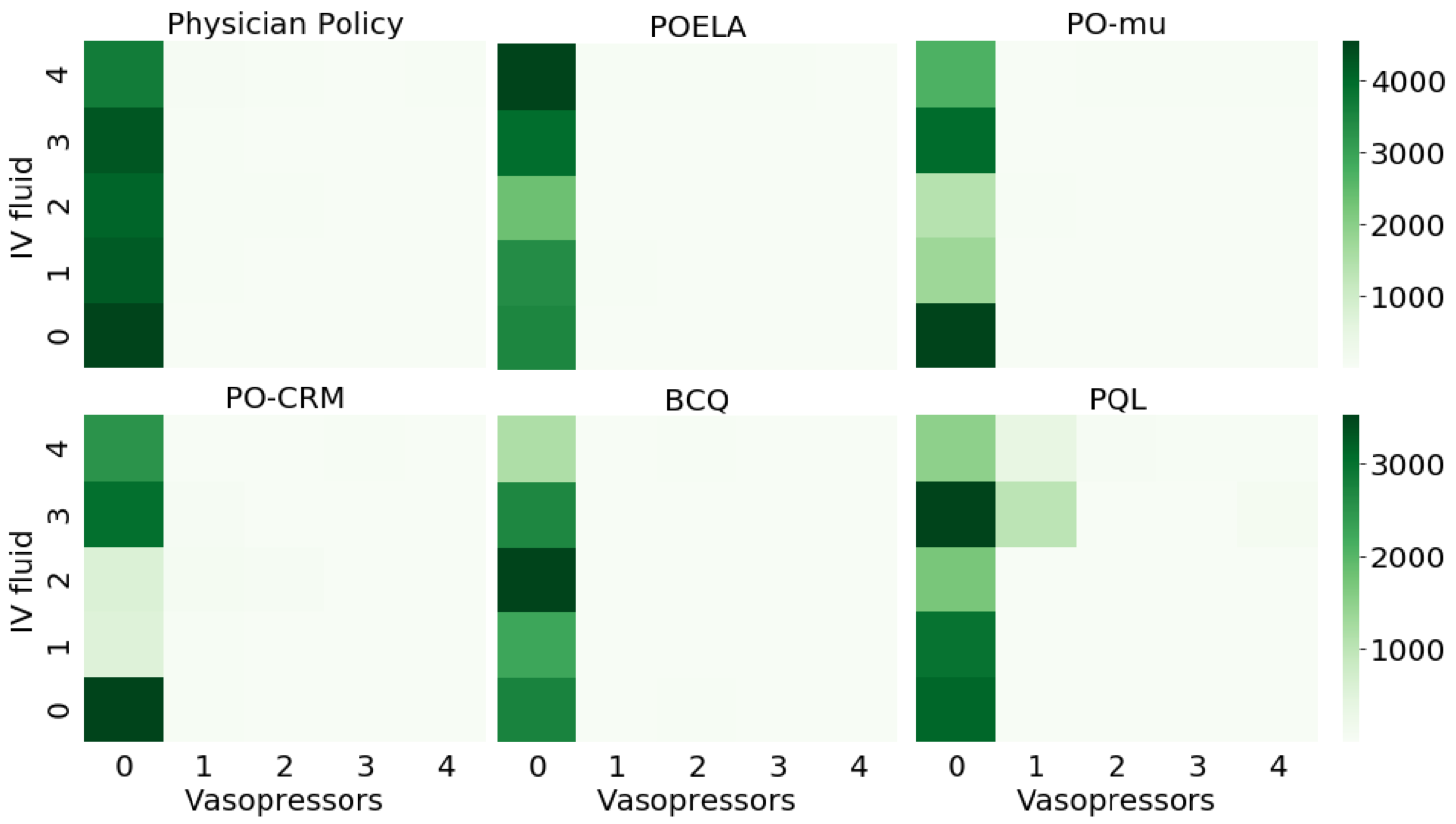}
    \subcaption{Action counts in low-SOFA contexts}
    \label{fig:sepsis_lowsofa_visualization}
    \end{minipage} 
    \caption{(a): Action counts heatmap in high-SOFA contexts of the policy learned from different algorithms. (b): Action counts heatmap in mid-SOFA contexts of the policy learned from different algorithms. (c): Action counts heatmap in low-SOFA contexts of the policy learned from different algorithms.}
\end{figure}

In this section, we explore the learned policies for patients with high logged SOFA scores (measuring organ failure) in the test dataset. Figure~\ref{fig:sepsis_highsofa_visualization} illustrates the number of actions taken by different policies and the clinicians. \ispg mainly takes treatments similar to the clinician's but more concentrated on high-vasopressors treatments, while \ispgbaseline and value-based methods take treatments different from the logged clinician decisions, suggesting these policies may be overfitting to avoid contexts with high SOFA. However, some patients arrive with high SOFA scores and a policy must have suitable treatments to support such individuals, which our method appears to ensure. For completeness, we also show the visualization of mid-SOFA ($5-15$) and low-SOFA ($<5$) patient contexts in Figures~\ref{fig:sepsis_midsofa_visualization} and~\ref{fig:sepsis_lowsofa_visualization}.

\clearpage
\section{Additional experiments: OpenAI Gym environment CartPole}
\label{ap:more-exp-cartpole}

In this experiment, we collect a dataset by training DQN~\citep{mnih2013playing} on the task and saving trajectories of horizon 200 steps at regular checkpoints during the training. The dataset is composed of a mixture of sub-optimal and expert data totalling 20000 transitions. For the non-Markov modification, we keep the \textit{Cart Position}, \textit{Cart Velocity} and \textit{Pole Angle} observations but remove the \textit{Pole Angular Velocity} element. In Table~\ref{tab:hp_cartpole}, we report the hyperparameter used in the experiments.

\begin{table}[ht]
    \centering
    \begin{tabular}{c|cc}
    \toprule
       Hyperparameters  & used in algorithms & values  \\
       \midrule
       $\delta$ & \ispg & $0.0001, 0.0005, 0.001, 0.005, 0.01$ \\
       $\hat{\mu}$ threshold & \ispgknn & 0.05, 0.1, 0.15, 0.2 \\
       CRM Var coefficient & \ispg, \ispgbaseline & $0, 0.1, 1, 10$ \\
       BCQ threshold & BCQ, PQL & $0.0, 0.05, 0.1, 0.2, 0.5$ \\ 
       \midrule
       $M$ in $\wtis$ & \text{All} & $1000$ \\
       Max training steps & \ispg, \ispgbaseline, \ispgknn & 500 \\
       & BCQ, PQL & 1000 \\
       Number of checkpoints & All & 50  \\
       Batch size & BCQ, PQL & 64 \\
       \bottomrule
    \end{tabular}
    \vspace{0.5cm}
    \caption{Hyperparameters in the CartPole experiment.}
    \label{tab:hp_cartpole}
\end{table}

\subsection{Standard evaluation procedure: use policy at the end of training}

\begin{table*}[ht]
    \centering
    \begin{small}
    \begin{tabular}{c|ccccc|c}
        \toprule 
         Method & \ispg& \ispgmuhat & \ispgbaseline & BCQ  & PQL & Behavior policy\\
         \midrule 
        Test SNTIS & 88.29 (86.62) & 78.79 & 72.63 & 21.28 & 23.61 & 41.41\\
        $95\%$ BCa UB & 89.70 (89.81) & 83.87 & 76.77 & 24.63 & 27.14 & 45.04\\
        $95\%$ BCa LB & 85.93 (85.57) & 69.64 & 68.15 & 16.22 & 20.36 & 38.16\\
        Test ESS & 43.32 (40.78) & 30.51 & 30.13 & 30.11 & 30.08 & 248\\
         \bottomrule
    \end{tabular}
    \end{small}
    \caption{CartPole dataset. Test evaluation, $(0.05, 0.95)$ BCa bootstrap interval, and ESS. The value of \ispg without a CRM variance penalty is shown in parentheses.}
    \label{tab:cartpole_results}
    \vspace{-0.3cm}
\end{table*}
\begin{table*}[ht]
    \centering
    \begin{small}
    \begin{tabular}{c|ccccc|c}
        \toprule 
         Method & \ispg& \ispgmuhat & \ispgbaseline & BCQ  & PQL & Behavior policy\\
         \midrule 
        Test SNTIS & 76.18 (72.21) &68.39 & 67.14 & 12.13 & 5.46 & 41.41\\
        $95\%$ BCa UB & 89.27 (88.32) & 80.22 & 83.72 & 12.89 & 6.63& 45.04\\
        $95\%$ BCa LB & 68.97  (67.49) & 57.13 & 57.78 & 9.17 & 5.02 & 38.16\\
        Test ESS & 36.41 (34.72) & 34.56 & 31.87 & 31.22 & 30.07 & 248\\
         \bottomrule
    \end{tabular}
    \end{small}
    \caption{Non-MDP CartPole dataset. Test evaluation, $(0.05, 0.95)$ BCa bootstrap interval, and ESS. The value of \ispg without a CRM variance penalty is shown in parentheses.}
    \label{tab:cartpole_pomdp_results}
    \vspace{-0.3cm}
\end{table*}

\subsection{Alternative selection procedure: checkpoint best intermittent policies}

\begin{table*}[ht]
    \centering
    \begin{small}
    \begin{tabular}{c|ccccc|c}
        \toprule 
         Method & \ispg& \ispgmuhat & \ispgbaseline & BCQ  & PQL & Behavior policy\\
         \midrule 
        Test SNTIS &  88.43 (87.56) & 76.01 & 82.25 & 17.74 & 17.83 & 41.41\\
        $95\%$ BCa UB & 90.46 (90.72) & 82.87 & 86.18 & 21.80 & 21.84 & 45.04\\
        $95\%$ BCa LB & 85.48 (84.63) & 66.21 & 74.30 & 12.84 & 13.26 & 38.16\\
        Test ESS & 43.32 (39.66) & 31.04 & 30.87 & 30.29 & 30.18 & 248\\
         \bottomrule
    \end{tabular}
    \end{small}
    \caption{CartPole dataset. Test evaluation, $(0.05, 0.95)$ BCa bootstrap interval, and ESS. The value of \ispg without a CRM variance penalty is shown in parentheses. \textbf{Procedure: best intermittent policy checkpoints.}}
    \label{tab:cartpole_results_checkpoint}
    \vspace{-0.3cm}
\end{table*}
\begin{table*}[ht]
    \centering
    \begin{small}
    \begin{tabular}{c|ccccc|c}
        \toprule 
         Method & \ispg& \ispgmuhat & \ispgbaseline & BCQ  & PQL & Behavior policy\\
         \midrule 
        Test SNTIS & 75.76 (75.70) &68.66 & 66.34 & 11.73 & 5.70 & 41.41\\
        $95\%$ BCa UB & 92.35 (89.16) & 79.56 & 82.46 & 12.49 & 6.71& 45.04\\
        $95\%$ BCa LB & 68.34  (66.08) & 55.49 & 57.50 & 7.98 & 5.08 & 38.16\\
        Test ESS & 37.72 (35.27) & 35.15 & 36.02 & 30.12 & 31.77 & 248\\
         \bottomrule
    \end{tabular}
    \end{small}
    \caption{Non-MDP CartPole dataset. Test evaluation, $(0.05, 0.95)$ BCa bootstrap interval, and ESS. The value of \ispg without a CRM variance penalty is shown in parentheses. \textbf{Procedure: best intermittent policy checkpoints.}}
    \label{tab:cartpole_pomdp_results_checkpoint}
    \vspace{-0.3cm}
\end{table*}

% \clearpage
\section{Additional experiments: D4RL}
\label{ap:more-exp-d4rl}
Although our primary focus is on application areas where the Markov assumption may not be correct or unverifiable, we also compare to an additional standard benchmark, namely D4RL.

An adaptation of the POELA algorithm is necessary to work with continuous action spaces. Practically, instead of using the eligible action set $A_h$, for each data sample, we pre-compute a set of similar actions and use the distance to the closest state $x_h$ associated with the most similar action distributions in the dataset as a smooth penalty in Line 5 of Algorithm~\ref{alg:is_policy_optimization}.

For each dataset quality (random, medium, and expert) and task (Hopper and Walker2D), we report the performances scaled from 0 to 100 (0 corresponds to the average returns of a random policy and 100 that of an expert policy) following the experimental protocol for D4RL with 200 episodes in each dataset. We compare with state-of-the-art methods in this dataset. The results are reported in Table~\ref{tab:d4rl_results}.

\begin{table*}[ht]
    \centering
    \begin{small}
    \begin{tabular}{c|ccc|c}
        \toprule 
         Dataset & \ispg & BCQ  & CQL & Behavior policy\\
         \midrule 
        Hopper-random & 10.5 & 10.5 & 10.8 & 9.8\\
        Hopper-medium & 43.7 & 42.9 & 41.4 & 29.0\\
        Hopper-expert & 58.9 & 59.7 & 52.6 & 43.6\\
        Walker2D-random & 6.1 & 4.6 & 5.4 & 1.6\\
        Walker2D-medium & 33.8 & 31.1 & 49.6 & 6.6\\
        Walker2D-expert & 32.2 & 32.8 & 54.7 & 50.2\\
         \bottomrule
    \end{tabular}
    \end{small}
    \caption{Additional experiments on 6 D4RL datasets.}
    \label{tab:d4rl_results}
    \vspace{-0.3cm}
\end{table*}

The results in Table~\ref{tab:d4rl_results} suggest that POELA performs similarly to two other state-of-the-art methods in this setting, even though POELA does not make Markov assumptions, which are made and leveraged in BCQ and CQL.

\end{document}